\documentclass[letterpaper, 10 pt, conference]{ieeeconf}  % Comment this line out if you need a4paper

\IEEEoverridecommandlockouts                              % This command is only needed if 

\usepackage[final]{graphicx}
\usepackage{hyperref}
\usepackage{amsmath}
\usepackage{amsfonts}
\usepackage{algorithm}
\usepackage{float}
\usepackage[caption = false]{subfig}
\usepackage{algpseudocode}
\graphicspath{{images/}}
\usepackage{mathtools} 

\usepackage[normalem]{ulem}

\usepackage[dvipsnames]{xcolor}

\begin{document}

%\runninghead{}

\title{\LARGE \bf A Safety-Aware Kinodynamic Architecture for Human-Robot Collaboration$^*$}

\author{Andrea Pupa$^1$, Mohammad Arrfou$^2$, Gildo Andreoni$^2$ and Cristian Secchi$^1$% <-this % stops a space
	\thanks{$^*$This project has received funding from the European Union’s Horizon 2020 research and innovation programme under grant agreement No. 818087 (ROSSINI).}% <-this % stops a space
	\thanks{$^{1}$ Andrea Pupa and Cristian Secchi are with the Department of Science and Method of Engineering, University of Modena and Reggio Emilia, Italy. E-mail:
		{\tt\small \{\href{mailto:andrea.pupa@unimore.it}{andrea.pupa}, \href{mailto:cristian.secchi@unimore.it}{cristian.secchi}\}@unimore.it}}%
	\thanks{$^{2}$ Mohammad Arrfou and Gildo Andreoni are with Datalogic S.p.A., Italy. E-mail:
		{\tt\small \{\href{mailto:mohammad.arrfou@datalogic.com.it}{mohammad.arrfou}, \href{mailto:gildo.andreoni@datalogic.com}{gildo.andreoni}\}@datalogic.com}}%%
}

%\keywords{}

\maketitle

\begin{abstract}
The new paradigm of human-robot collaboration has led to the creation of shared work environments in which humans and robots work in close contact with each other. Consequently, the safety regulations have been updated addressing these new scenarios. The mere application of these regulations may lead to a very inefficient behavior of the robot. In order to preserve safety for the human operators and allow the robot to reach a desired configuration in a safe and efficient way, a two layers architecture for trajectory planning and scaling is proposed. The first layer calculates the nominal trajectory and continuously adapts it based on the human behavior. The second layer, which explicitly considers the safety regulations, scales the robot velocity and requests for a new trajectory if the robot speed drops. The proposed architecture is experimentally validated on a Pilz PRBT manipulator.\\
\end{abstract}

% \begin{keywords}
%     Human-Robot Collaboration, robot safety, dynamic planning.
% \end{keywords}

\section{Introduction} 
% \begin{itemize}
% \item Human robot colaboration and safety in collaborative scenario
% \item Safety problem with consequent great limitation of the speed/approccio conservativo dell'umano
% \item State of the art on safety in HRC
% \item State of the art on trajectory scaling
% \item Importance of planning strategy in high dynamic environment
% \item State of the art of planner and replanner for robot manipulator
% \item Main contribution: two layers framework that allows to plan and replan an admissible trajectory and following the path respecting safety constraint
% \item Novelty with respect to the state of the art
% \end{itemize}
The introduction and diffusion of collaborative robotics within the industrial environments has allowed to create shared workspace where humans and robots can work closely. While this new paradigm has led to an increase in the flexibility of production lines, the lack of physical barriers requires to pay more attention on how to guarantee human safety. Therefore, the robot safety standards have been updated to address this new collaborative scenarios~\cite{villani2018}. In particular, the ISO 10218-1 and the ISO 10218-2~\mbox{\cite{iso2011robot-1,iso2011robot-2}} standards classify the collaborative modes in four different categories: \textit{safety-rated monitored stop} (SMS), \textit{hand guiding} (HG), \textit{speed and separation monitoring} (SSM) and \textit{power and force limiting} (PFL). Additionally, the technical specification ISO/TS 15066 \cite{isots} provides further information to assess the risk for each collaboration mode. In case of applications where industrial robots are used, the SSM is typically adopted. In this collaborative mode the speed of the robot is reduced according to the relative human-robot velocity and position. However, this approach is overly conservative, since the robot speed should not be limited if its motion is directed away from the human. Moreover, by monitoring the human speed the performance of the robot can be further increased without violating the safety constraints.

Different approaches were presented in the literature to deal with human safety and collision avoidance in a human-robot collaboration (HRC) scenario. 
In \cite{ragaglia2015} the authors propose a real-time solution to evaluate the future human occupancy and scale the robot speed accordingly, ensuring safety. The idea is to use a 3D camera and a simple human kinematic model to predict the future human occupancy.
In \cite{zanchettin2015} an optimization which treats safety as an hard constraint to be satisfied is presented. This strategy leads to obtain a proportional reduction of the speed, with a consequent higher productivity, while ensuring safety.
In \cite{lippi2020}  the authors  present a safety framework for collaborative tasks where multiple robots have to share the workspace with human operators. The idea is to scale the velocity preventing that a safety index falls below a certain value. When the scaling procedure is not enough, an emergency stop is applied.

Reducing the speed of the robot is not always the best solution, especially when the workspace conditions allow the robot to modify the pre-planned path. In \cite{levratti2019} the authors exploit the concept of static and kinetostatic danger field on a mobile robot in order to prevent collision with human operators in a tire workshop. In \cite{chen2018} the concept of potential field around the whole robot body is used to generate collision-free trajectory. The entire workspace is surrounded by multiple depth sensors that track both dynamic and static object.  In \cite{ferraguti2015} authors implement virtual fixtures, which combine attractive and repulsive potential field, in a teleoperated environment. Even if these methods are effective in guaranteeing safety requirements, potential fields can easily cause the system to be stuck in local minima, compromising the task execution.

For this reason, optimization-based algorithms have been exploited to achieve a collision free behavior by applying the minimum correction to the desired path. Safety is embedded through the constraints in the optimization problem. In \cite{lin2017} an optimization problem is solved in real-time in order to force the robot to stay inside a safe set, evaluating the variation of a safety index. 
% In \cite{ferraguti2020}, the authors propose an optimization-based control algorithm to avoid the human operator while trying to preserve the desired path. Their strategy exploits the use of Control Barrier Functions \cite{ames2016} around the whole robot body to maintain a collision-free trajectory.
In \cite{ferraguti2020-iso}, the authors propose an optimization-based control algorithm that explicitly considers safety in order to avoid the human operator while trying to preserve the desired path. Their strategy exploits the use of control barrier functions \cite{ames2016} around the robot body to maintain a collision-free trajectory while fulfilling the ISO/TS 15066.

Adopting the optimal behavior to avoid collision in highly dynamic environments could be computationally challenging, especially in a real industrial scenario where the number of obstacles to be considered is very high. In \cite{lavalle2001, kunz2014} the authors use kinodynamic rapidly-exploring random tree~(RRT) to plan collision free trajectory under kinodynamic constraints. However, these solutions are only suitable for constraints that do not change during the execution of the path, while the safety kinodynamic constraints change in real time based on human behavior.

Other solutions, like \cite{udai2014}, propose to ensure safety by making the robot behave like a passively compliant system during the execution of a task. In spite of that, these approaches treat the human operator as an external disturb for the system, without exploiting a human tracking strategy.

In this paper we propose a novel framework for trajectory planning and velocity scaling for HRC scenario that is aware of the highly dynamic of the environment and ensures safety for the human operator by explicitly considering safety regulations.
% In this paper we propose a novel framework for trajectory planning and velocity scaling for HRC application that explicitly considers safety regulations ensuring safety for the human operator. Moreover, in order to avoid a drastic drop of the robot speed with consequent, 
The proposed framework is composed by two layers.
Given a desired configuration to reach, a trajectory planner layer computes and adapts online the trajectory that the robot has to follow. The trajectory scaling layer, according to the safety constraints imposed by the safety standards, scales the robot velocity ensuring safety for the human operator. Moreover, in order to avoid drastic drops of the robot velocity with consequent poorly efficient robot behaviors, mutual communication between the two layers is enabled. When required, the trajectory scaling can request for a replan of a new trajectory, increasing the robot performances.

The main contributions of this paper are:
\begin{itemize}
	\item A novel adaptive framework for trajectory planning and scaling that takes into account the high dynamicity of the environment, adapting in real-time the trajectory. 
	\item A strategy for trajectory scaling that is computationally cheap, i.e. suitable for real industrial application, and that explicitly considers the kindoynamic safety constraint.
	\item The overall architecture that integrates the trajectory planning and scaling strategies in order to improve the efficiency of the system.
\end{itemize}

The paper is organized as follows: in Sec.~\ref{sec:problem} the trajectory planning and scaling problem is detailed while in Sec.~\ref{sec:ssm} the SSM collaborative mode is treated. In Sec.~\ref{sec:architecture} the overall architecture is presented: in Sec.~\ref{subsec:dynamic_planner} the trajectory planning strategy is detailed, while Sec~\ref{subsec:task_scaling} the trajectory scaling problem considering safety constraints is presented. Finally in Sec.~\ref{sec:experiments} an experimental validation of the proposed architecture is presented while in Sec.~\ref{sec:conclusions} some conclusions and future works are addressed.
\section{Problem Statement} 
\label{sec:problem}
% Trattare i due problemi contemporaneamente
% \begin{itemize}
%     \item The trajectory scaling problem
%     \begin{itemize}
%         \item points on the robot
%         \item points on the human
%         \item limitation of the velocity against the human
%     \end{itemize}
% \item The planner and replanner strategy
%   \begin{itemize}
%       \item RRT-Connect
%       \item not optimal
%       \item very fast and suitable for highly dynamic environments
%   \end{itemize}
% \end{itemize}
We consider a HRC application where a robot manipulator with $n$ joints has to move from an initial configuration $q(t_i) = q_i \in \mathbb{R}^n$ to a desired final configuration \mbox{$q(t_f) = q_f \in \mathbb{R}^n$} in order to execute a task. The trajectory $q(t) \in \mathbb{R}^n$ that the robot has to perform can be decomposed with a path-velocity decomposition:
\begin{equation}
    \label{eq:q_s}
    q(t) = q(s(t))
\end{equation}
where $s$ is the curvilinear abscissa that parametrizes the geometrical path $q(s)$. The variation of $s$ represents the time law of the desired path (i.e. the velocity profile).

Differentiating \eqref{eq:q_s} we obtain:
\begin{equation}
    \label{eq:q_s_dot}
    \dot{q}(t) = q'(s)\dot{s}
\end{equation}
where $q'(s)$ is the vector tangent to the desired path, while $\dot{s}$ constitutes the magnitude of the joint velocity.

The trajectory $q(t)$ is considerate feasible and collision-free when:
\begin{equation}
    \begin{split}
        d(\sigma_{ri}(\bar{q}),\sigma_{Hj})\ge d_{min} \quad &\forall i \in \{1,\dots,n\}, \forall \bar{q} \in q(t),\\&\forall j \in \{1,\dots,m\}
    \end{split}
\end{equation}
where $\sigma_{ri}(\bar{q})$ is the line segment representing the $i$-th link when the robot is in configuration $\bar{q}$. $\sigma_{Hj}$ it the line segment of the $j$-th human body link, e.g. the human arm, and $m$ is the number of the human body link. $d(\sigma_{ri}(\bar{q}),\sigma_{Hj})$ represents the distance between the two line segments and $d_{min}$ is the minimum admissible distance.
For this reason, the shared workspace is equipped with a monitoring system that allows to track the human movements and estimate the human speed. Several strategies to track the human body are available in literature: skeleton tracking with multiple cameras \cite{moon2016},  placing markers on the human body \cite{kofman2005}, machine learning techniques \cite{fan2010}, to name a few.

In this work, we aim at designing a safety kinodynamic architecture that:
\begin{itemize}
    \item Computes a nominal trajectory that is always collision-free, i.e. a trajectory that the robot can execute at maximum speed. Exploiting the tracking of the human movements, the planning strategy aims at preserving the feasibility of the trajectory, replanning a new trajectory when the actual trajectory becomes infeasible.
    \item Starting from the nominal trajectory, scales the robot velocity according to the limits imposed by the ISO/TS 15066 standard. The scaling aims at maintaining safety for the human operator taking into account both the distance between human and robot and the velocity of the human towards the robot.
\end{itemize}

\section{Speed and Separation Monitoring} 
\label{sec:ssm}
\begin{figure}[t]
	\centerline{\includegraphics[width=0.7\columnwidth]{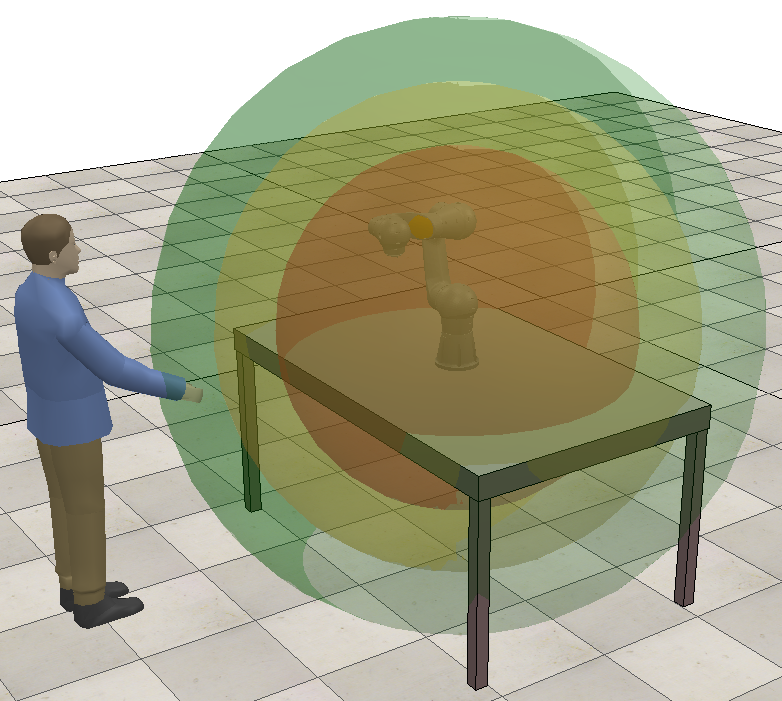}}
	\caption{Representation of different safety zones with SSM collaboration mode.} 
	\label{fig:ssm}
\end{figure}

% \begin{itemize}
%     \item normative resume
%     \item type of HRC
%     \item Speed and separation monitoring
%     \item Collaborative job
%     \item Using SSM also in collaborative scenario
% \end{itemize}
% As a consequence of the development of HRC technologies, the safety standards have been updated in order to address the new co-working scenarios. In particular, the ISO 10218-1/2 standard classify the collaborative modes in four different categories: \textit{Safety-rated Monitored Stop} (SMS), \textit{Hand Guiding} (HG), \textit{Speed and Separation Monitoring} (SSM) and \textit{Power and Force Limiting} (PFL). In case of applications where industrial robots are used, the Speed and Separation Monitoring is typically used. 
In modern industrial applications of collaborative robotics, the Speed and Separation Monitoring collaboration mode is widely used.
In this collaborative mode, the speed of the robot is continuously adapted depending on the position and velocity of the human operator into the collaborative workspace. Typically the human velocity is not monitored and the workspace is divided into three different areas based on the distance between the human and the robot. This scenario is represented in Fig.~\ref{fig:ssm}. The robot is allowed to operate at full speed when the human is in the green area, at reduced speed when the human is in the yellow area and it stops when the human is in the red area.
The ISO/TS 15066 provides the guidelines for calculating the sizes of these areas, namely the minimum protective separation distance $S_p$, considering also the relative speed between the robot and the human operator. $S_p$ can be computed as:
\begin{equation}
    \label{eq:minimumdistance}
    S_p(t_0) = S_h + S_r + S_s + C + Z_d + Z_r
\end{equation}
$S_p$ is the protective separation distance at time $t_0$, while $t_0$ is the current time. $S_h$ represents the contribution to the protective separation distance due to the operator's movements, $S_r$ is the one derived from the robot reaction time and $S_s$ is the contribution caused by the robot stopping time. $C$ represents the intrusion distance, i.e. the distance that a part of the body can intrude into the sensing field before it is detected. $Z_d$ and $Z_r$ are the position uncertainties of the human operator inside the workspace and of the robot system respectively.

The first terms of \eqref{eq:minimumdistance} can be expressed as:
\begin{align}
    \label{eq:contribution1}
    S_h &= \int_{t_0}^{t_0+T_s+T_r}v_h(t)\,dt
    \\\nonumber\\
    \label{eq:contribution2}
    S_r &= \int_{t_0}^{t_0+T_r}v_r(t)\,dt
    \\\nonumber\\
    S_s &= \int_{t_0+T_r}^{t_0+T_s+T_r}v_s(t)\,dt \label{eq:contribution3}
 %=
    % \\
    % &=\int_{t_0+T_r}^{t_0+T_s+T_r}v_r(t)\,dt + a_{max}\frac{(T_s+T_r)^2}{2}
\end{align}
where $T_s$ and $T_r$ represents the robot stopping time and the robot reaction time respectively. $v_h$ is the directed speed of the human operator towards the robot, $v_r$ is the directed speed of the robot towards the human operator and $v_s$ is the speed of the robot in the course of stopping.
%$a_{max}$, instead, is the maximum deceleration of the robot in the stopping phase.

Under the assumptions that the velocity of the robot is constant during the robot reaction time, that the acceleration remains constant during the stopping phase and that the dynamics of the human operator is slower than the robot dynamics, which is true in the case of a generic HRC application, the equations \eqref{eq:contribution1}~--~\eqref{eq:contribution3} can be approximated as follow:
\begin{align}
    \label{eq:contributionel1}
    &S_h = v_h(t_0)(T_s+T_r)
    \\\nonumber\\
    \label{eq:contributionel2}
    &S_r = v_r(t_0)T_r
    \\\nonumber\\
    \label{eq:contributionel3}
    &S_s = v_r(t_0)T_s+a_{max}\frac{T_s^2}{2}
\end{align}

Substituting \eqref{eq:contributionel1}~--~\eqref{eq:contributionel3} in \eqref{eq:minimumdistance}, it is possible to obtain an upper bound robot velocity:
\begin{align}
    \nonumber v_{r_{max}}(t_0)=&\frac{S_p(t_0)-v_h(t_0)(T_s+T_r) - C - Z_d - Z_r}{T_s+T_r}
    \\
    \label{eq:vellimit}
    &- \frac{a_{max}T_s^2}{2(T_s+T_r)}
\end{align}
% \begin{equation}
%     \scalebox{0.9}{$v_r(t_0)=2\frac{S_p(t_0)-v_h(t_0)(T_s+T_r)+ C + Z_d + Z_r}{2T_r+T_s} - \frac{a_{max}(T_s+T_r)^2}{2T_r+T_s}$}
% %    \scalebox{1}{$v_r(t_0)=\frac{S_p(t_0)-v_h(t_0)(T_s+T_r)-\frac{a_{max}(T_s+T_r)^2}{2}+ C + Z_d + Z_r}{2T_r+T_s}$} 
% \end{equation}

The equation \eqref{eq:vellimit} is the safety constraint imposed by the ISO/TS 15066, i.e. it expresses the maximum speed allowed to the robot in the direction of the human operator.

\section{Safety Kinodynamic Architecture}
\label{sec:architecture}
The proposed dynamic trajectory planning and scaling strategy can be represented by the architecture in Fig.~\ref{fig:framework}, where two main layers can be distinguished:
\begin{figure}[t]
\vspace*{5pt} 
	\centerline{\includegraphics[width=0.85\columnwidth]{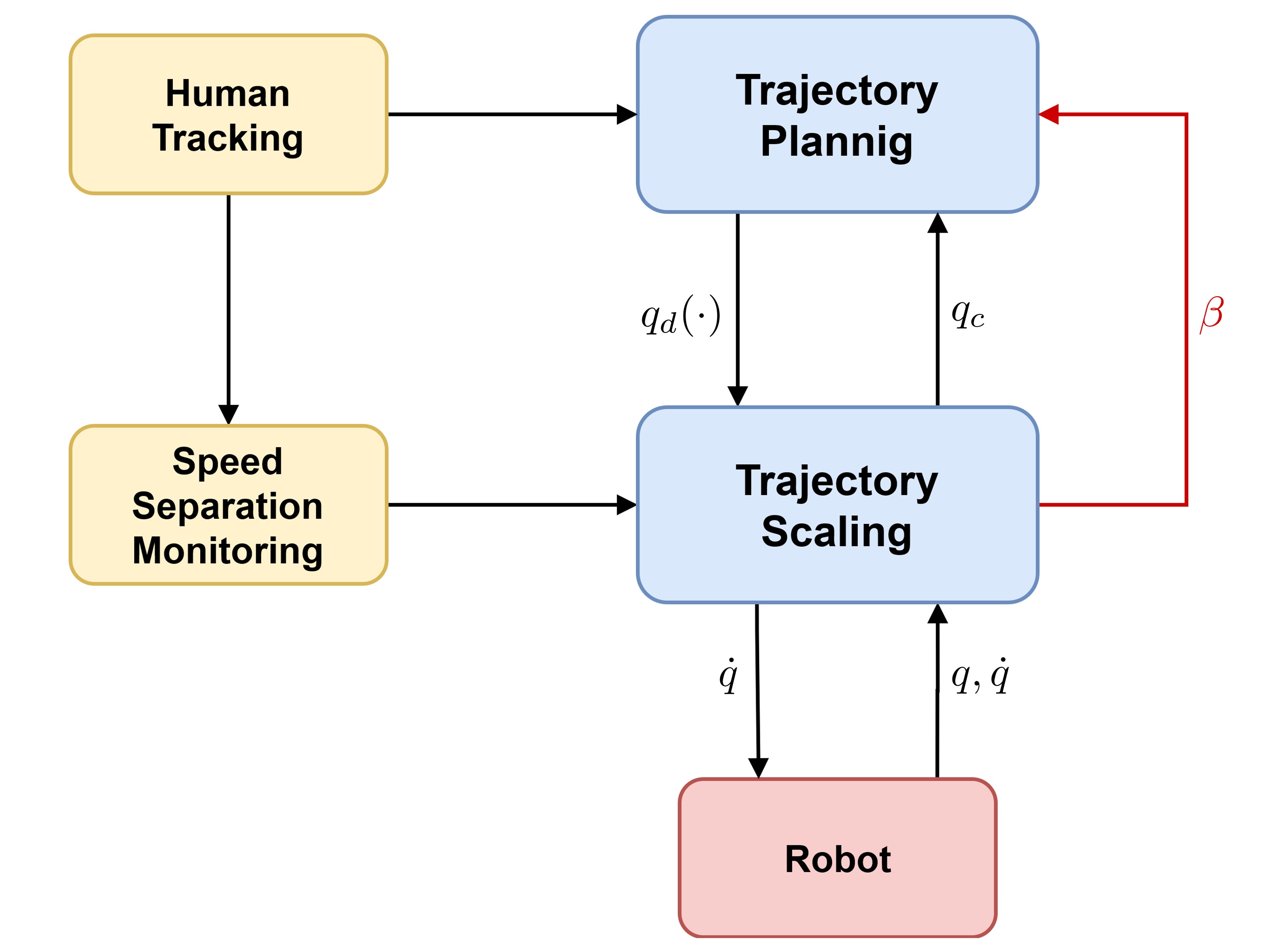}}
	\caption[caption]{The overall architecture. The blue blocks represent the two layers. The yellow blocks, instead, symbolize the strategies implemented to provide richer information to the layers. The red block represents the agent.\newline The black lines symbolize the data exchange, while the red one constitutes the signal that request for a replan of a new trajectory.} 
	\label{fig:framework}
\end{figure}
\begin{enumerate}
	\item \textbf{The trajectory planning layer}. It is responsible of generating the initial nominal trajectory that the robot can execute at maximum speed, i.e. it considers only the robot limits. Subsequently, it continuously adapts this trajectory exploiting the human tracking information.
	\item \textbf{The trajectory scaling layer}. It is responsible of scaling the robot velocity along the planned path, explicitly taking into account the velocity limits imposed by the safety constraints.
\end{enumerate}

Once the trajectory planning computes the initial nominal trajectory, it sends it to the trajectory scaling and it remains active until the robot reaches the desired final configuration $q_f$. The trajectory planning layers does not take into account the safety regulation, i.e. it computes a trajectory that the robot could ideally execute at maximum speed. 

The trajectory scaling firstly applies a path-velocity decomposition to the desired trajectory as shown in \eqref{eq:q_s} and \eqref{eq:q_s_dot}. Subsequently, it computes online the optimal scaled velocities in order to satisfy the constraint imposed by ISO/TS 15066 \eqref{eq:vellimit}.

During the execution of the motion, mutual communication between the two layers is enabled. The trajectory planning exploits the human tracking information and replans a new trajectory when the previous one becomes infeasible, as explained in Sec.~\ref{subsec:dynamic_planner}.
The trajectory scaling immediately parameterizes the new trajectory and starts following the new path. At each iteration, it returns to the trajectory planning the actual state of the trajectory. Moreover, when the scaling factor decreases too much, the trajectory scaling sends a signal to the trajectory planning requesting for a new trajectory to be planned, as it is becoming inefficient, see Sec.~\ref{subsec:task_scaling}.

It is worth noting that during the real-time execution the two algorithms work in parallel, relying on the last available data sent by the other algorithm. For an optimal behavior, the scaling algorithm should work at a frequency at most equal to that of the robot control.
\subsection{Trajectory Planning}
\label{subsec:dynamic_planner}
% \begin{itemize}
%     \item RRT-Connect
%     \item using in a dynamic environment
%     \item pseudocode of the dynamic planner
% \end{itemize}
The role of this layer is to find a trajectory $q_d(t)$ for the robot that is collision-free and that the robot can execute at maximum speed. Since the human behavior is in general unpredictable it is not possible to use a strategy that computes offline an optimal trajectory, as in short time it could become infeasible causing collisions between the human and the robot. The trajectory planning aims at continuously maintaining a collision-free trajectory, adapting it online when required.

The trajectory planning is implemented according to the pseudo-code reported in Alg.~\ref{alg:dynamicplanner}.
\begin{algorithm}
    \caption{TrajcetoryPlanning()}
    \label{alg:dynamicplanner}
    \begin{algorithmic}[1]
    	\State \textbf{Require:} $q_i,q_f,n$\label{algl:dprequire}
    	\State $q_d(\cdot)\leftarrow\mathbf{plan}(q_i, q_f)$ \label{algl:dpplan}
    	\State$\mathbf{send}(q_d(\cdot))$\label{algl:dpsend}
    	\State$q_c\leftarrow q_i$\label{algl:dpinit}
    	\While{$q_c \ne q_f$ \label{algl:dpwhiles}}
    	\State$h\leftarrow\mathbf{horizon}(q_c,n)$ \label{algl::dphorizon}
    	\For{$i=1:n$}\label{algl:dpforloop}
    	\If{$\mathbf{not\_feasible}(h(i))$ \label{algl:dpcheck}}
    	\State $q_d(\cdot)\leftarrow \mathbf{replan}(q_d(\cdot),q_d(h(i-1)),q_f)$\label{algl:dpreplaninfeasible}
    	\State \textbf{break}
    	\EndIf
    	\EndFor
    	\If{$\beta$}\label{algl:dpread}
    	\State $q_d(\cdot)\leftarrow \mathbf{replan}(q_d(\cdot),q_c,q_f)$\label{algl:dpreplanineff}
    	\EndIf
    	\State $\mathbf{update\_q_c}()$\label{algl:dpupdate}
    	\EndWhile
    % 	\While{$s_r<s_{end}$\label{algl:dpwhiles}}
    % 	\State $(feasible, s_{free}) \leftarrow\mathbf{check}(q_d(s),s_r)$\label{algl:dpcheck}
    % 	\If{$\mathbf{not} feasible$\label{algl:dpiffeasible}}
    % 	\State $q_d(s)\leftarrow \mathbf{replan}(q_d(s),s_{free},q_f)$\label{algl:dpreplaninfeasible}
    % 	\EndIf
    % 	\State $q_{int}\leftarrow\mathbf{read\_request}()$ \label{algl:dpread}
    % 	\If{$q_{int} \ne \emptyset$ \label{algl:dpif}}
    % 	\State $q_d(s)\leftarrow\mathbf{plan}(q_d(s_r), q_{int})$ \label{algl:dpplanint}
    % 	\State $q_d(s)\leftarrow \mathbf{replan}(q_d(s),s_{end}, q_f)$\label{algl:dpreplanint}
    % 	\EndIf
    % 	\State $s_r\leftarrow \mathbf{update\_s}()$\label{algl:dpupdate}
    % 	\EndWhile
    \end{algorithmic}
\end{algorithm}

The trajectory planning needs as input the initial and the final configuration, respectively $q_i$ and $q_f$, and the length of the horizon trajectory that will be checked $n$ (Line~\ref{algl:dprequire}). It immediately plans the maximum speed trajectory $q_d(\cdot)$ that the robot could perform (Line~\ref{algl:dpplan}). The function $\mathbf{plan}$ can be implemented using different strategies available for robotic applications (see e.g. \cite{lavalle1998, jaillet2004, ratliff2009}). Subsequently it sends the trajectory to the trajectory scaling layer (Line~\ref{algl:dpsend}) and it sets the current trajectory state $q_c$ equal to the initial configuration $q_i$ (Line~\ref{algl:dpinit}). From this point the algorithm starts to loop until the entire trajectory has been executed (Line~\ref{algl:dpwhiles}). In the loop, the dynamic planner first creates the horizon $h$ starting from the actual state (Line~\ref{algl::dphorizon}). This horizon represents the set of the future configuration that are analyzed to check if the trajectory is still feasible (Line~\ref{algl:dpforloop}~--~\ref{algl:dpcheck}). In case an infeasible configuration is found  a new feasible trajectory is planned through the function $\mathbf{replan}$ (Line~\ref{algl:dpreplaninfeasible}). The $\mathbf{replan}$ function is responsible of planning a new trajectory that goes from a desired configuration, in this case the last feasible one $q(h(i-1))$, to the final goal. Moreover, the $\mathbf{replan}$ function merges the new trajectory with the previous one and sends the resulting trajectory to the trajectory scaling. Subsequently, the dynamic planning algorithm reads if there is a request to replan a new trajectory due to the inefficiency of the current one, i.e. $\beta$ is equal to one (Line~\ref{algl:dpread}). This request is given by trajectory scaling layer, as described in Sec.~\ref{subsec:task_scaling}. If there is the request, a new trajectory starting from the actual configuration is computed (Line~\ref{algl:dpreplanineff}). Lastly, the actual configuration is updated exploiting the information coming from the trajectory scaling in \eqref{eq:qc} (Line~\ref{algl:dpupdate}).

The replan algorithm is presented in Alg.~\ref{alg:replan}.
\begin{algorithm}
    \caption{replan()}
    \label{alg:replan}
    \begin{algorithmic}[1]
    	\State \textbf{Require:} $q_d(\cdot),q_{rp}, q_f$\label{algl:rrequire}
    	\State $q_{new}(\cdot)\leftarrow\mathbf{plan}(q_{rp}, q_f)$ \label{algl:rplan}
    	\State $q_d(\cdot)\leftarrow\mathbf{merge}(q_d(\cdot), q_{new}(\cdot))$\label{algl:rmerge}
    	\State$\mathbf{send}(q_d(\cdot))$\label{algl:rsend}
    	\State$\mathbf{return} (q_d(\cdot))$\label{algl:rreturn}
    \end{algorithmic}
\end{algorithm}	

The algorithm takes as input the actual planned trajectory $q_d(\cdot)$, the starting configuration of the new trajectory $q_{rp}$ and the final desired configuration $q_f$ (Line~\ref{algl:rrequire}). It firstly plan a new trajectory $q_{new}(\cdot)$ that goes from the starting configuration of the new trajectory $q_{rp}$ to the desired goal $q_f$ (Line~\ref{algl:rplan}). The new trajectory is then merged with the old one  (Line~\ref{algl:rmerge}). This merging procedure replaces the part of the old trajectory from $q_{rp}$ to $q_f$ with the new trajectory. Lastly, the updated trajectory $q_d(\cdot)$ is sent to the trajectory scaling (Line~\ref{algl:rsend}) and returned to the dynamic planner (Line~\ref{algl:rreturn}).
\subsection{Trajectory Scaling}
\label{subsec:task_scaling}
% \begin{itemize}
%     \item task scaling problem
%     \item very detailed explanation of the constraints
%     \item modified jacobian
% \end{itemize}
Starting from the output of the dynamic planner, the goal of the trajectory scaling is to regulate the robot velocity without violating the safety constraint expressed in \eqref{eq:vellimit}. When a human and a robot cooperate the environment could be highly dynamic, for this reason the robot must follow exactly the same path coming from the upper layer, since a deviation from the planned path could cause a collision. The trajectory scaling aims at scaling only the magnitude of the velocity $\dot{s}$, assuring that the executed path is collision-free.

This is achieved in two steps. Firstly, by applying the path-velocity decomposition as shown in \eqref{eq:q_s}~--~\eqref{eq:q_s_dot}. Secondly, by solving the following optimization problem:

\begin{equation}
    \label{eq:problem}
    \begin{array}[l]{ll}
        min-\alpha \\\\
        \text{subject to}\\\\
        J_{r_i}({q})q'_d(s)\dot{s}\alpha \leq V_{max} \quad \forall i \in \{1,\dots,n\}\\\\
        \dot{q}_{min} \leq q'_d(s)\dot{s}\alpha \leq \dot{q}_{max}\\\\
        \ddot{q}_{min}\leq \frac{q'_d(s)\dot{s}\alpha - \dot{q}}{T_r} \leq \ddot{q}_{max}\\\\
        0 \leq \alpha \leq 1
        % \alpha \in
        % \begin{bmatrix}
        % 0,1
        % \end{bmatrix}
    \end{array}
\end{equation}
$\alpha \in \begin{bmatrix}0, 1 \end{bmatrix}$ is the optimization variable and represents the scaling factor.  $J_{r_i}(q)\in \mathbb{R}^{1 \times n}$ is a \textit{modified jacobian} that takes into account only the scalar velocity towards the human operator of the $i$-th link. This modified version of the jacobian is required as the velocity constraint imposed by the ISO/TS 15066 \eqref{eq:vellimit} limits only the velocity that reduce the human-robot distance, i.e. the velocity towards the human. $V_{max} \in \mathbb{R}^{n}$ is a vector whose each component is the velocity limit imposed by the ISO/TS 15066. $\dot{q}_{min} \in \mathbb{R}^{n}$ and $\dot{q}_{max} \in \mathbb{R}^{n}$ are the joint velocity lower bounds and the joint velocity upper bounds, respectively. While $\ddot{q}_{min} \in \mathbb{R}^{n}$ and $\ddot{q}_{max} \in \mathbb{R}^{n}$ are the acceleration limits. $\dot{q} \in \mathbb{R}^{n}$ is the actual robot velocity and $T_r$ is the robot execution time.

The modified jacobian $J_{r_i}(q)$ is expressed as:
\begin{equation}
    J_{r_i}(q) = \Vec{n}_i^T\,
    \begin{bmatrix}
    J_i(q)\quad
    \bar{0}
    \end{bmatrix}
\end{equation}
where $\Vec{n}_i = \{n_{x_i}, n_{y_i}, n_{z_i}, 0, 0, 0\}$ is the versor representing the direction that goes from the $i$-th robot link to the human.The method used to compute this versor is a design parameter, e.g. it can be found representing both the robot and the human links as capsules and computing the minimum distance \cite{ferraguti2020}. $J_i(q) \in \mathbb{R}^{6 \times i}$ is the $i$-th jacobian, i.e. the jacobian matrix that relates the firsts $i$ joint velocity to the linear and angular velocity of the $i$-th link, and $\bar{0}\in \mathbb{R}^{6 \times (n-i)}$ is a matrix with all zero elements. 
% , i.e $J_i(q)$ is the Jacobian that considers only the first $i$ joints. In order to maintain the same dimensions, $n-i$ empty columns are added.

The optimization problem \eqref{eq:problem} is a convex problem and computationally cheap, since the only factor that affects the convergence is the problem dimension, i.e. the number of joints and links. Thanks to its convexity, the solution obtained by the solver is always the global minimum of the cost function, i.e. the maximum admissible scaling factor. Moreover the problem has always a feasible solution. When the human operator is very far from the robot, the robot is allowed to move at the desired speed, i.e. $\alpha = 1$ that is the maximum speed as seen in Sec.~\ref{subsec:dynamic_planner}. When the human approaches the robot, the safety standards require to decrease the velocity until, in the worst case, stopping the robot. This is guaranteed by the solution $\alpha = 0$.

The output of the trajectory scaling is then used to send the desired velocity to the robot:
\begin{equation}
    \dot{q} = q'_d(s)\dot{s}\alpha 
\end{equation}
while the new robot configuration that is sent to the dynamic planner (see Alg.\ref{alg:dynamicplanner}, Line~\ref{algl:dpupdate}) will be:
\begin{equation}
    \label{eq:qc}
    q_c = q_c+q'_d(s)\dot{s}\alpha T_r
\end{equation}

However, greatly reduce the robot velocity is a very conservative strategy and it strictly decreases the overall efficiency. Sometimes it could be more convenient for the robot to move away from the human and execute another trajectory. For this reason it has been implemented a step signal that requests to the dynamic planner the replan of new trajectory:
\begin{equation}
    \beta = \begin{cases} 
      1 & \alpha\leq \alpha_{min} \\
      0 & otherwise
   \end{cases}
\end{equation} 

where $\alpha_{min}$ is a predefined threshold and represents the lower desired bound for the scaling factor.

% To be sure that the robot can reach high speeds, the new trajectory should go, at least for some time, in the opposite direction to that towards the human. \ap{An intermediate point $P_{int}$ that the robot has to reach before going towards the goal is then added}. This new point is defined as:
% \begin{equation}
%     P_{int} = -S_p\begin{bmatrix}n_x\\ n_y\\ n_z  \end{bmatrix}
% \end{equation}
% The intermediate point is then transformed to a configuration in the joint space exploiting the inverse kinematics $IK$:
% \begin{equation}
%     q_{int} = IK(P_{int})
% \end{equation}

% \ap{Once $q_{int}$ is calculated a replan request is sent to the Dynamic Planner and a new trajectory is planned (see Alg.~\ref{alg:dynamicplanner}, Line \ref{algl:dpread}-\ref{algl:dpreplanint}).}
When $\beta$ is high a replan request is sent to the trajectory planning and a new trajectory is planned (see Alg.~\ref{alg:dynamicplanner}, Line~\ref{algl:dpread}).
\section{Experiments} 
\label{sec:experiments}
\begin{figure}[t]
\vspace*{5pt} 
	\centerline{\includegraphics[width=0.8\columnwidth]{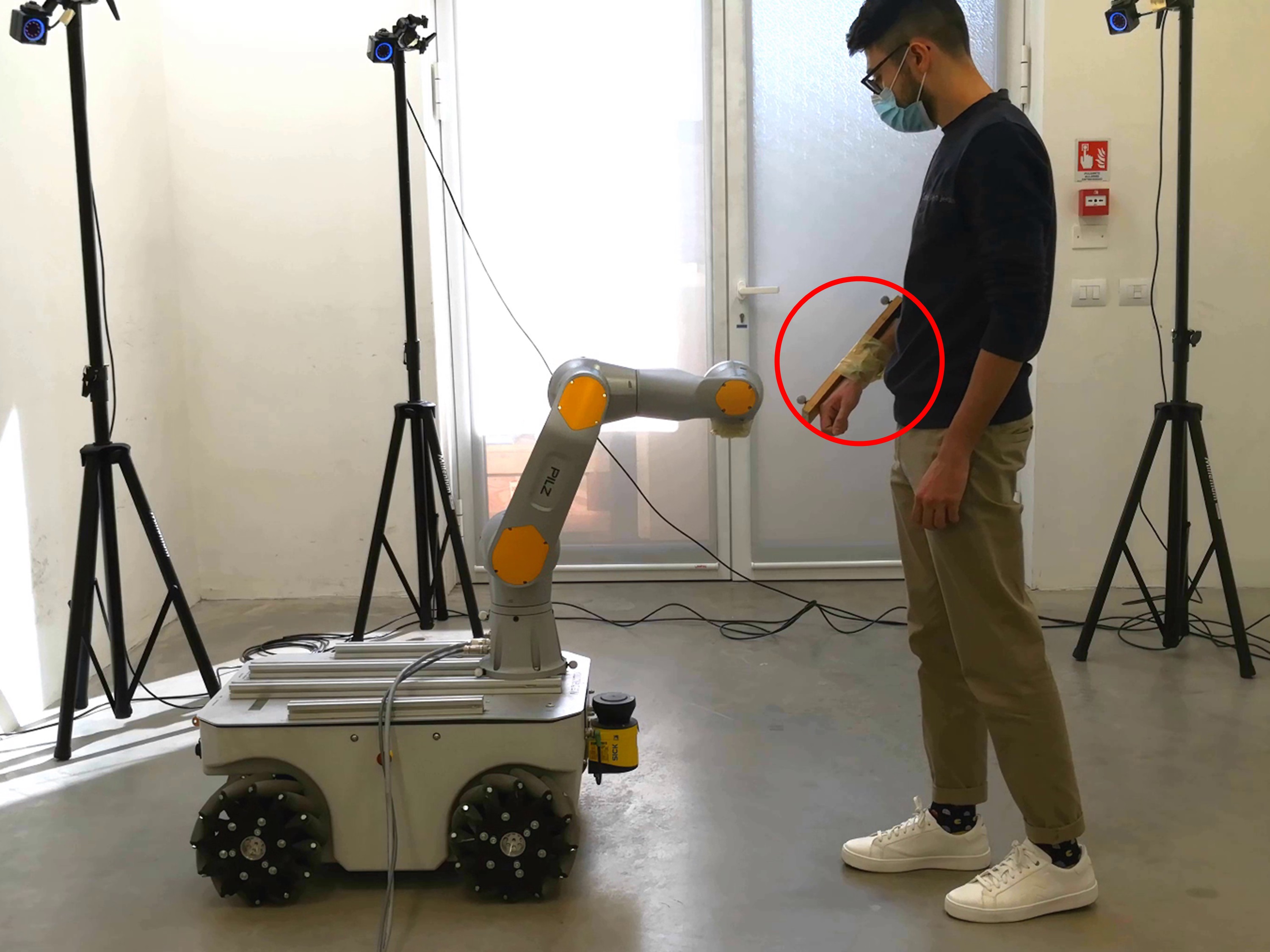}}
	\caption{Setup of the experiments. The Pilz PRBT manipulator, which is placed on a mobile robot, three of the six OptiTrack Prime$^\text{x}$ cameras and a wooden rod with the OptiTrack markers to track the right arm of the human operator (red circle).} 
	\label{fig:setup}
\end{figure}

% \begin{itemize}
%     \item Description of the experiment
%     \item Hardware and algorithms used
%     \item Experimental results
%     \item Discussion
%     \item Scenario robot does something at maximum speed (maybe pick and place) and then when the huan approaches it start going very slow. The human put the arm inside the trajectory and the robot avoid it.
% \end{itemize}
The proposed two-layers framework has been experimentally validated on a Pilz PRBT, a 6-DoF manipulator for industrial application. We decided to exploit six OptiTrack Prime$^\text{x}$ cameras with the OptiTrack Motive software \cite{optitrack} in order to track the movements of the human right arm.
A complete setup of the experiments is shown in Fig.~\ref{fig:setup}.
All the software components were developed using ROS Melodic Morenia meta-operating system and they ran on a Intel(R) Core(TM) i7-10510U with Ubuntu 18.04. The dynamic planner layer is based on the RRT-Connect algorithm~\cite{rrt-connect} and it is implemented using MoveIt Motion Planning Framework~\cite{moveit}. The trajectory scaling layer exploits the C code generated by CVXGEN \cite{cvxgen} to solve the optimization problem \eqref{eq:problem}. For simplicity, the modified jacobian $J_{r_i}$ is applied only to the end-effector and the versor $\Vec{n}_i$ is the versor of minimum distance between the $i-th$ robot link and the human operator arm. The minimum distance is computed representing both the robot links and the human arm as capsules (see \cite{ferraguti2020}).

Concerning the frequencies, the communication with the robot works at $50\,\,Hz$ while the optimization problem converges in $1\,\,ms$. The OptiTrack, instead, works at a frequency of $240\,\,Hz$. Since the PRBT has not a real-time velocity ROS interface, it has been decided to position control the robot integrating the solution coming from \eqref{eq:problem} at $20\,\,Hz$.

\begin{figure}[t]
    \subfloat[Nominal Positions]{\includegraphics[width=0.5\columnwidth]{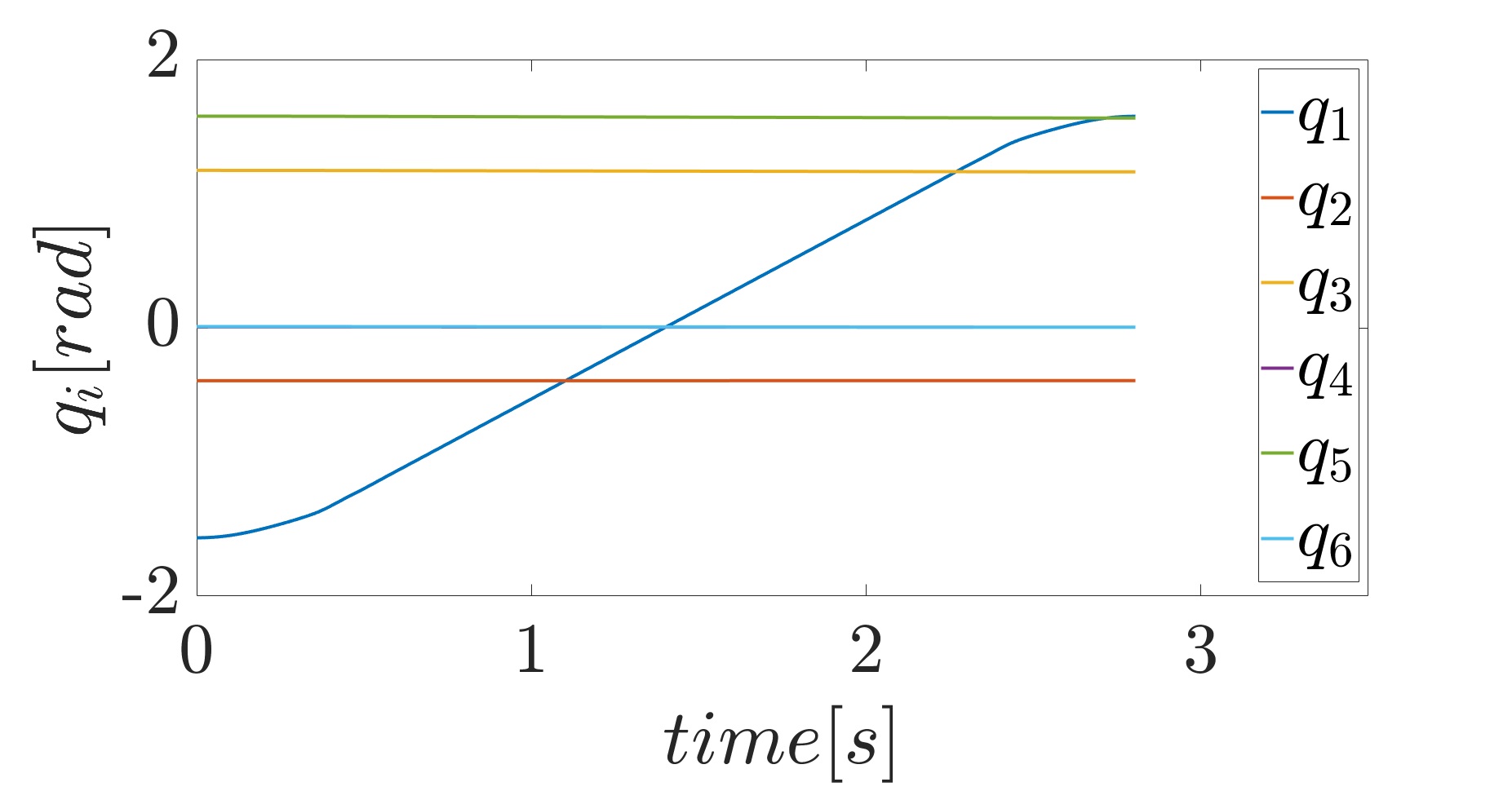}}
    \subfloat[Real Positions]{\includegraphics[width=0.5\columnwidth]{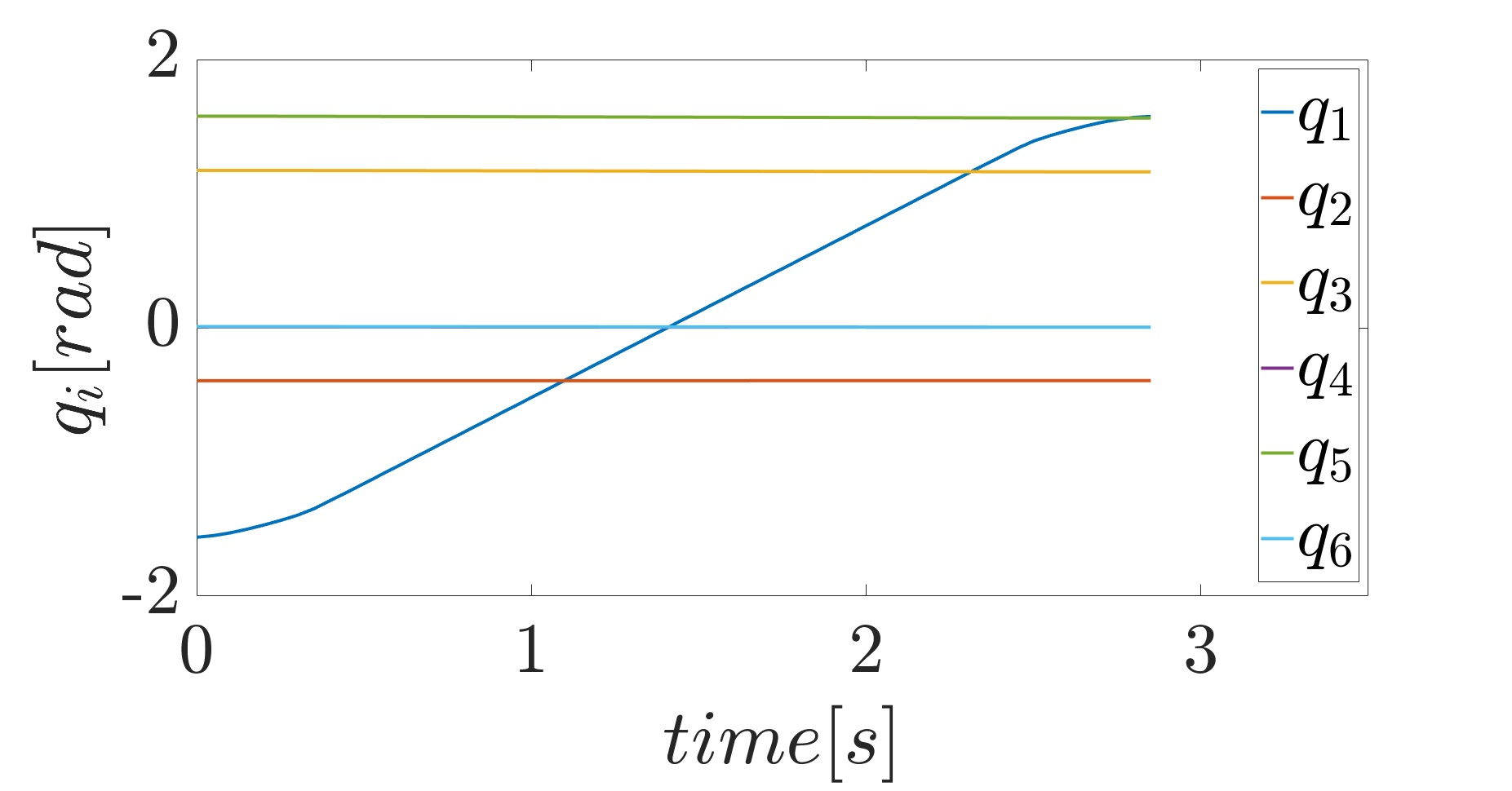}}\\
    \subfloat[Nominal Velocities]{\includegraphics[width=0.5\columnwidth]{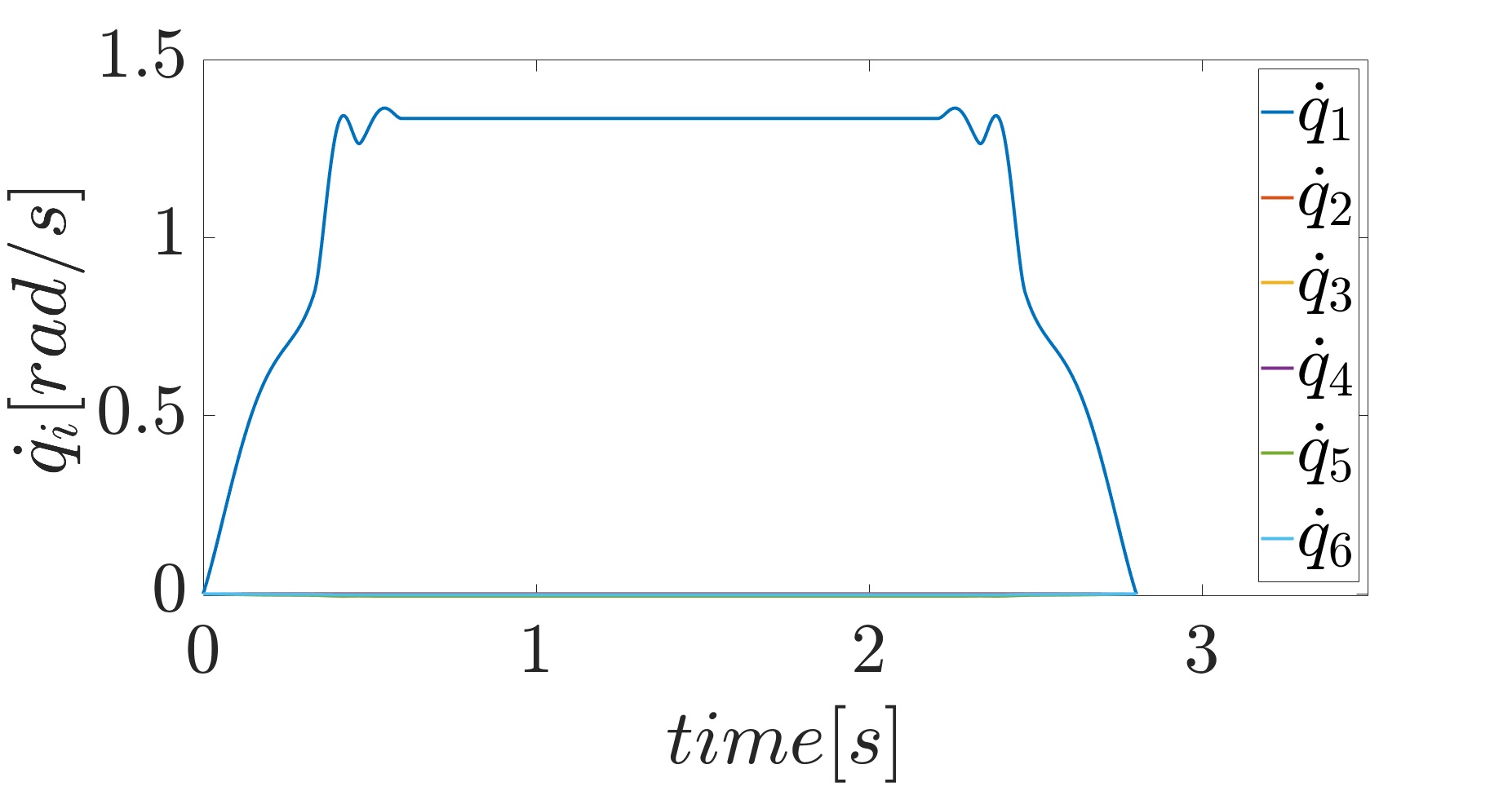}}
    \subfloat[Real Velocities]{\includegraphics[width=0.5\columnwidth]{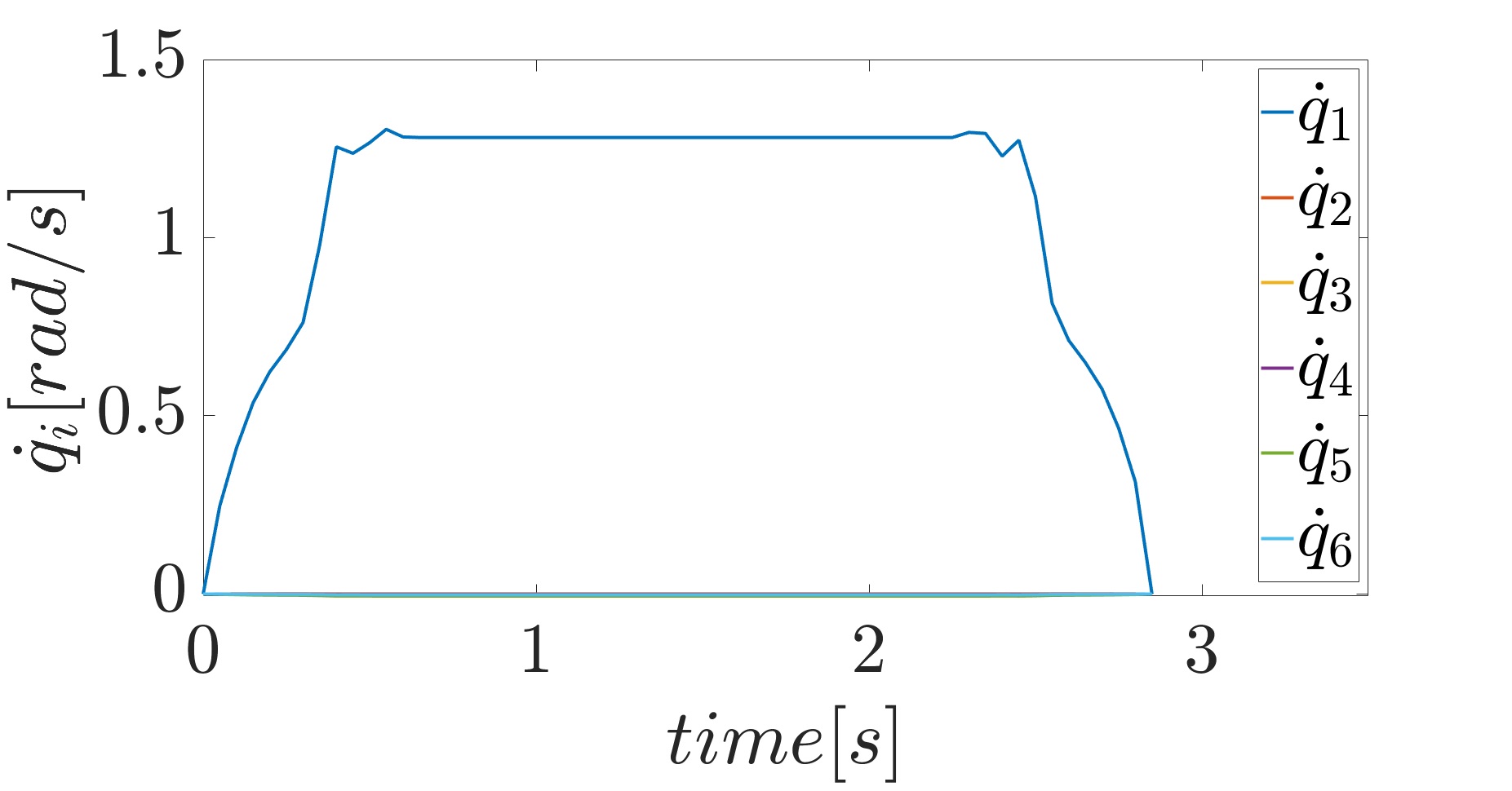}} 
    \caption{First part of the experiment. Since the human operator is far from the robot, the robot velocity is not scaled.}
    \label{fig:nominal}
\end{figure}

\begin{figure}[t]
\subfloat[Human Position]{\includegraphics[width=0.5\columnwidth]{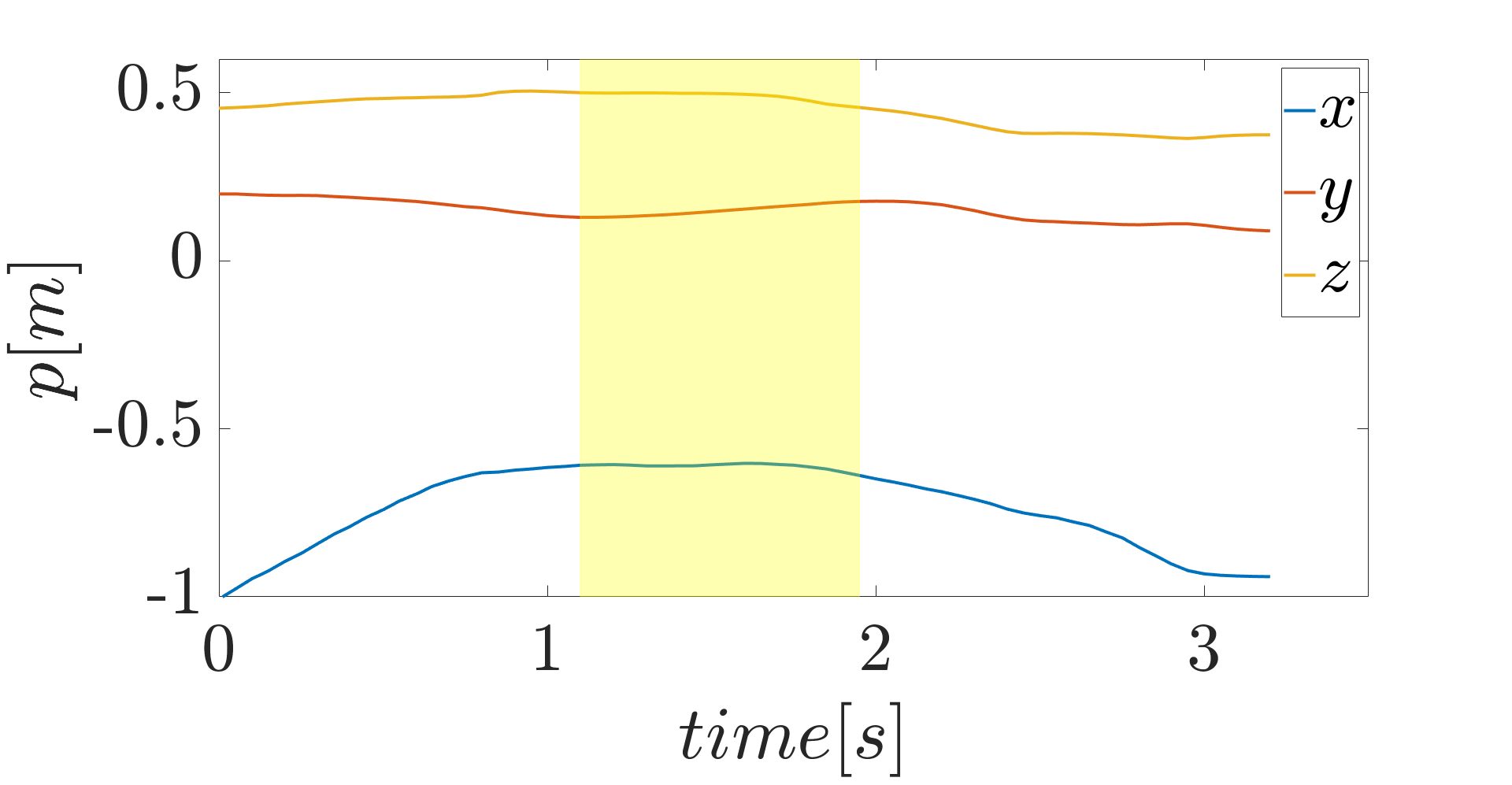}\label{fig:human_pose}}
        \subfloat[Human Velocity]{\includegraphics[width=0.5\columnwidth]{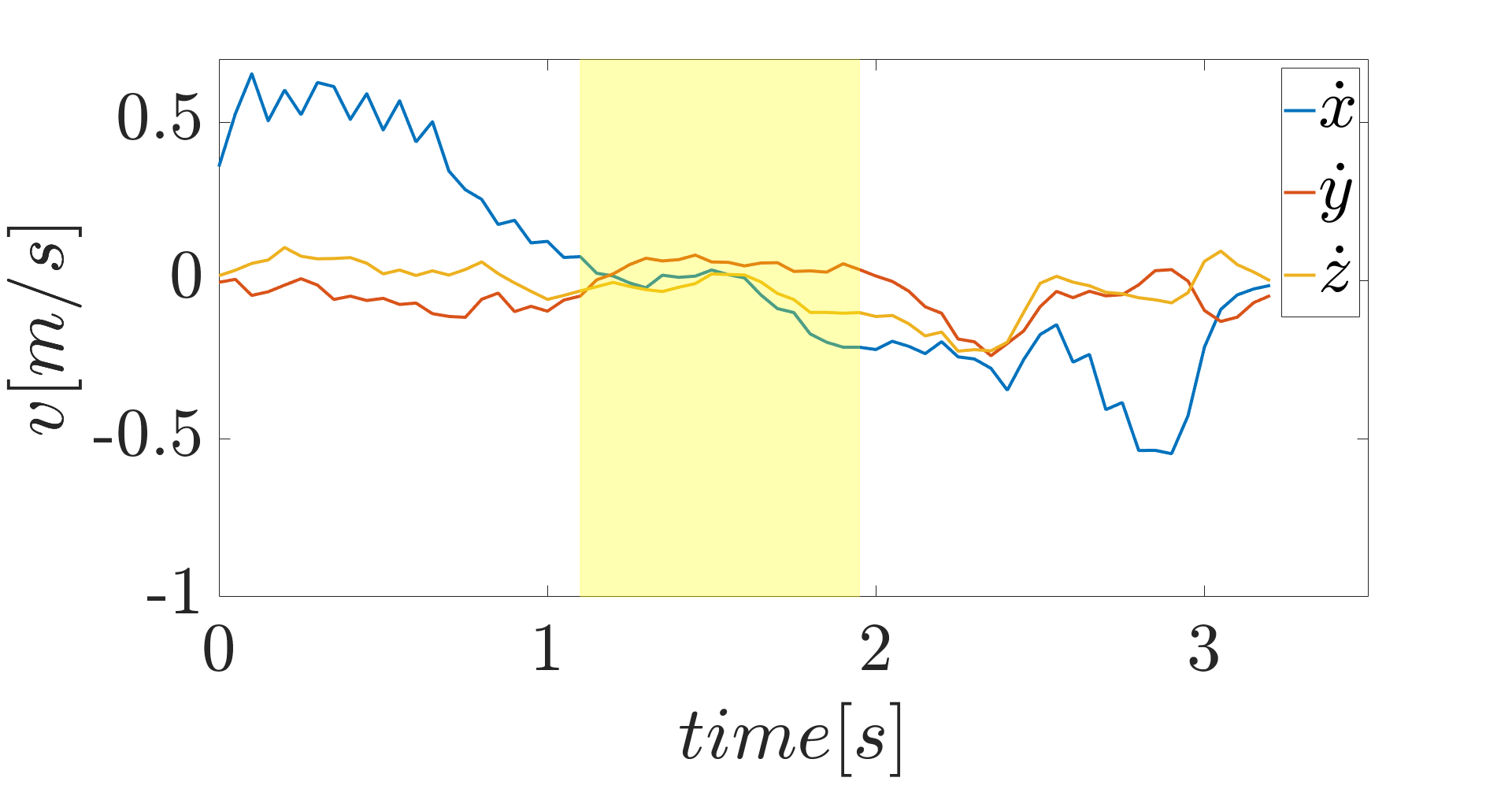}\label{fig:human_vel}}\\
    \subfloat[Scaling Factor and Step Signal]{\includegraphics[width=0.5\columnwidth]{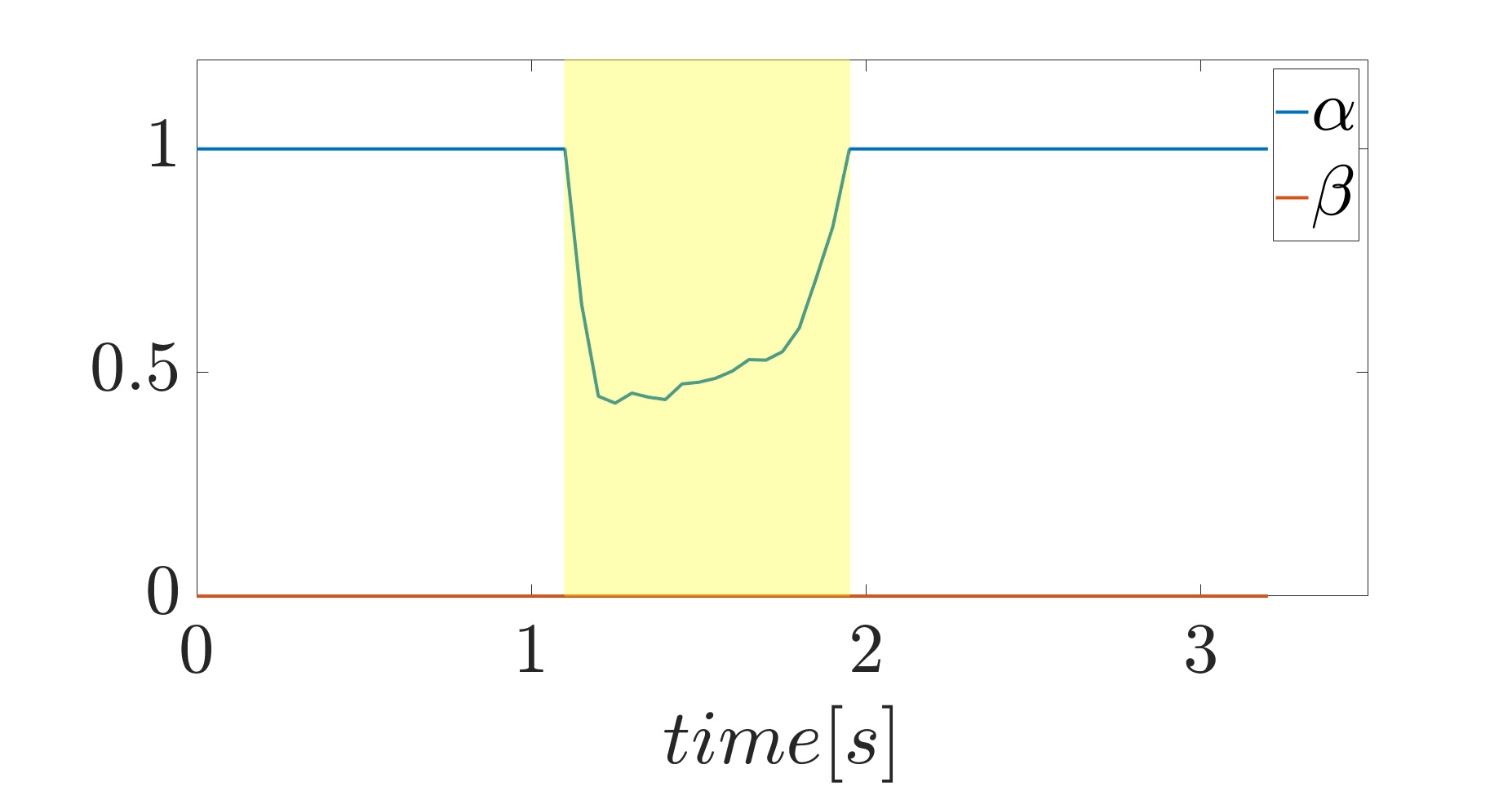}\label{fig:scale}}
    \subfloat[Velocity Constraint on the EE]{\includegraphics[width=0.5\columnwidth]{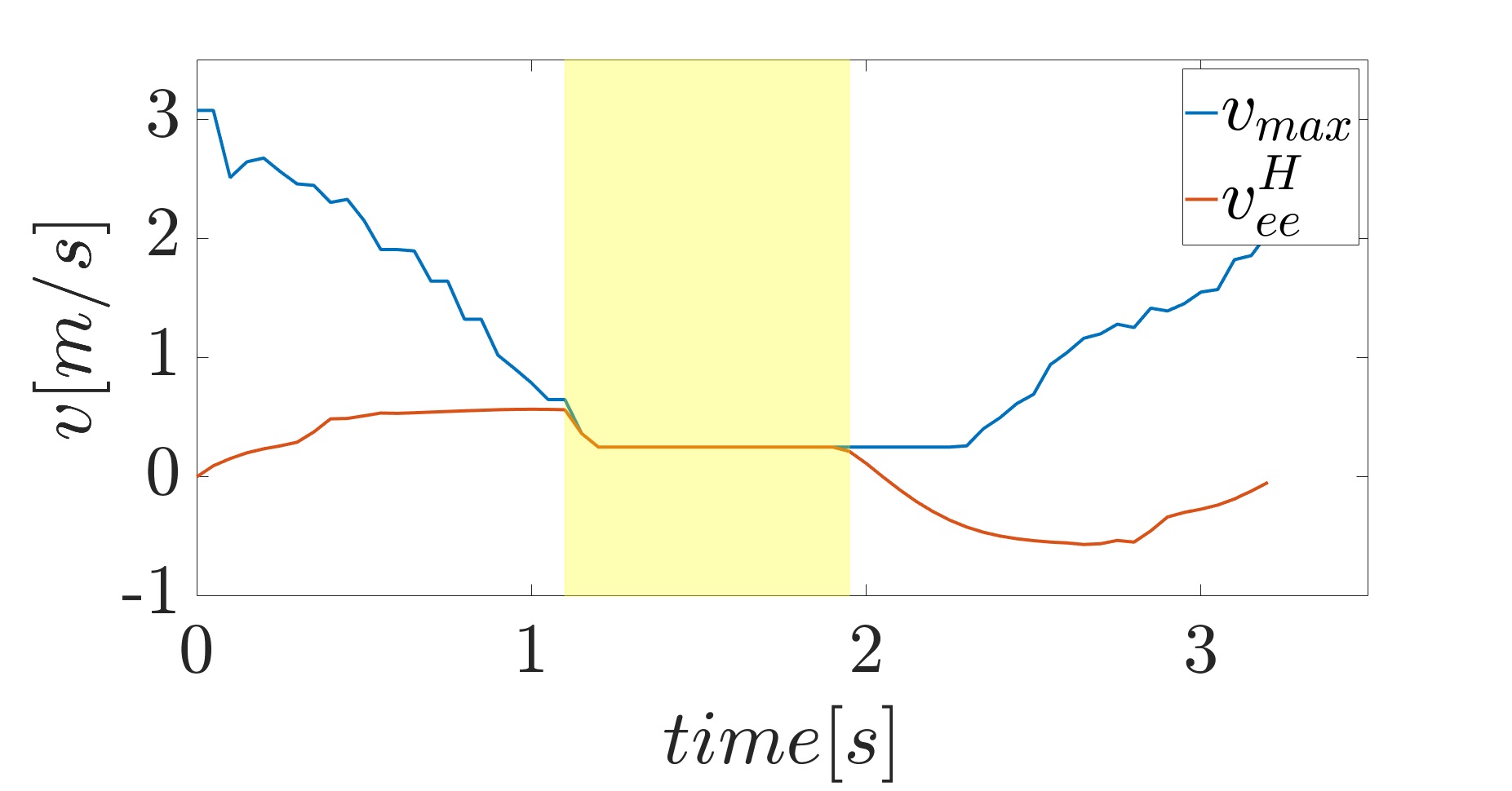}\label{fig:velconst}}
        \caption{Second part of the experiment. The human operator approaches the robot and the velocity is scaled in order to fulfill the safety constraint.} 
\end{figure}

\begin{figure}[t]
    \subfloat[Nominal Positions]{\includegraphics[width=0.5\columnwidth]{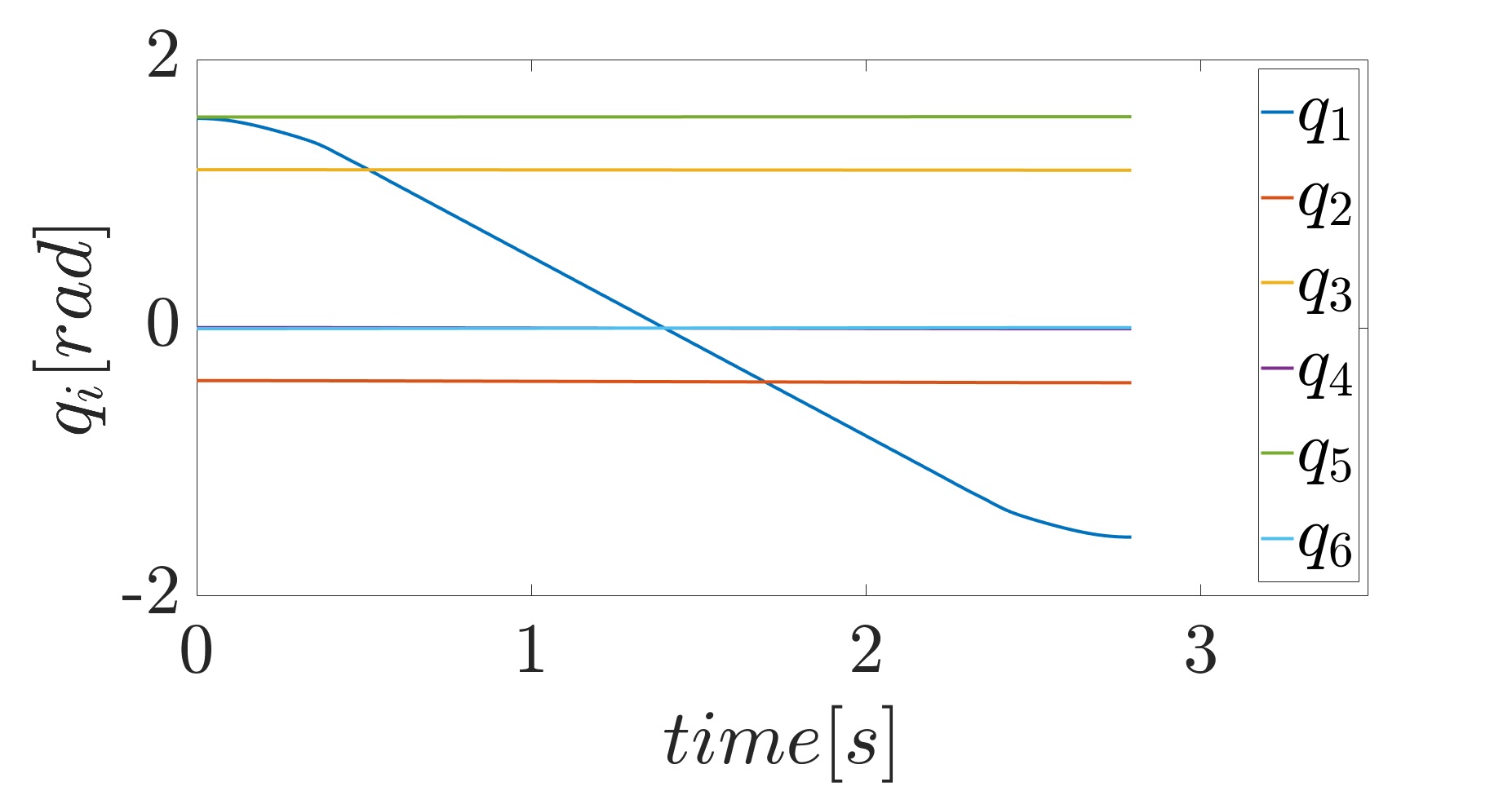}}
    \subfloat[Real Positions]{\includegraphics[width=0.5\columnwidth]{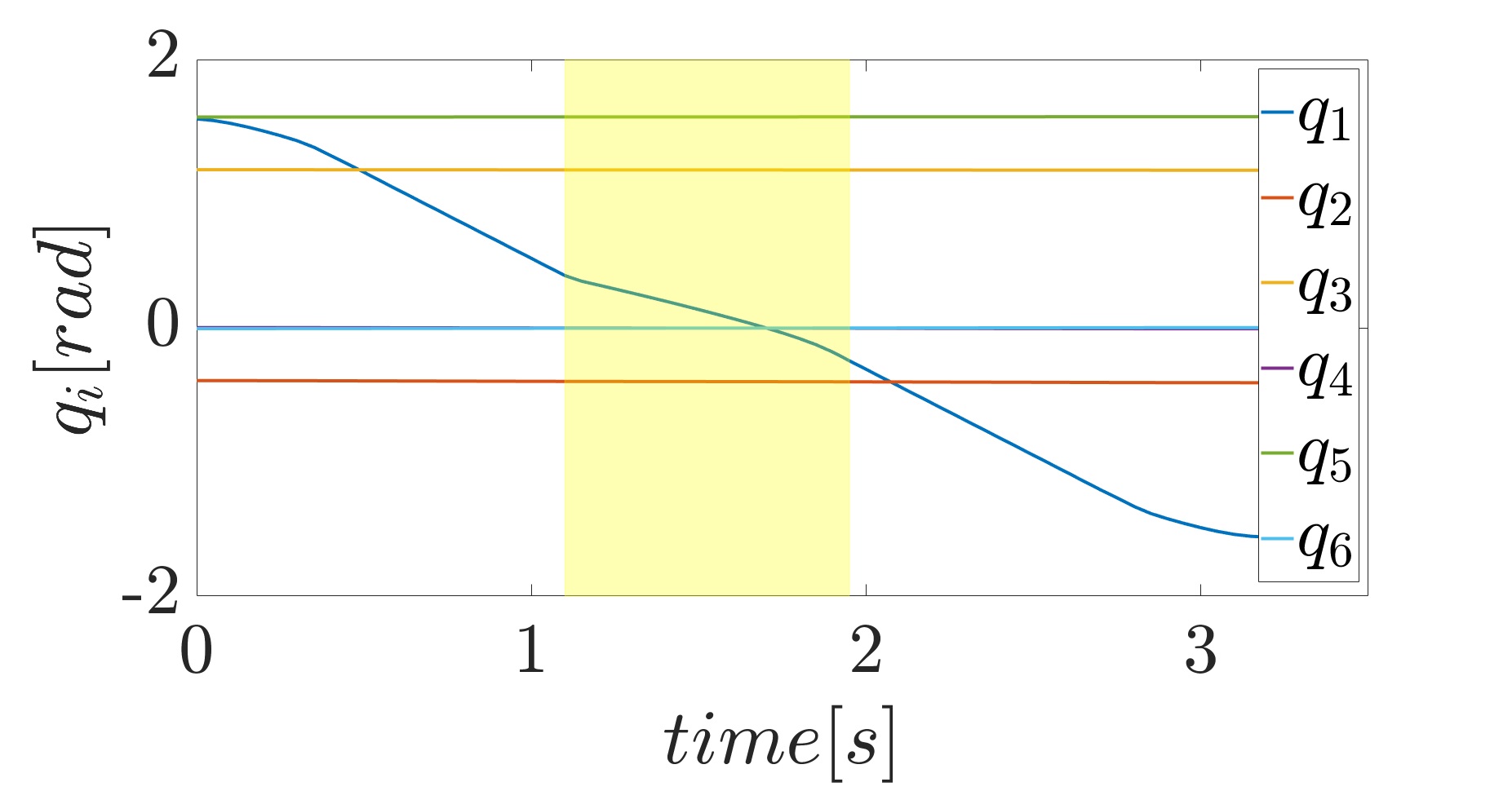}}\\
    \subfloat[Nominal Velocities]{\includegraphics[width=0.5\columnwidth]{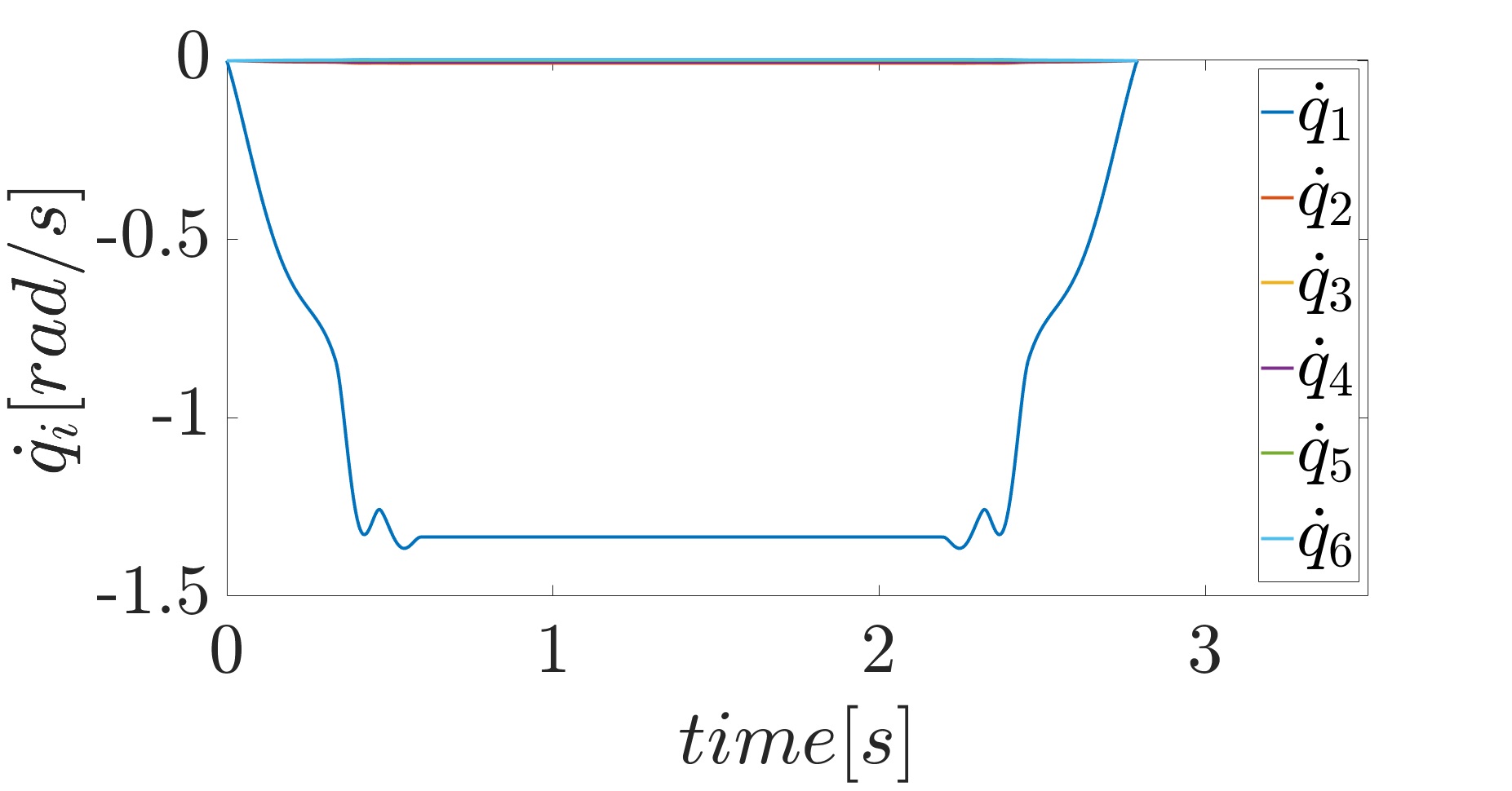}}
    \subfloat[Real Velocities]{\includegraphics[width=0.5\columnwidth]{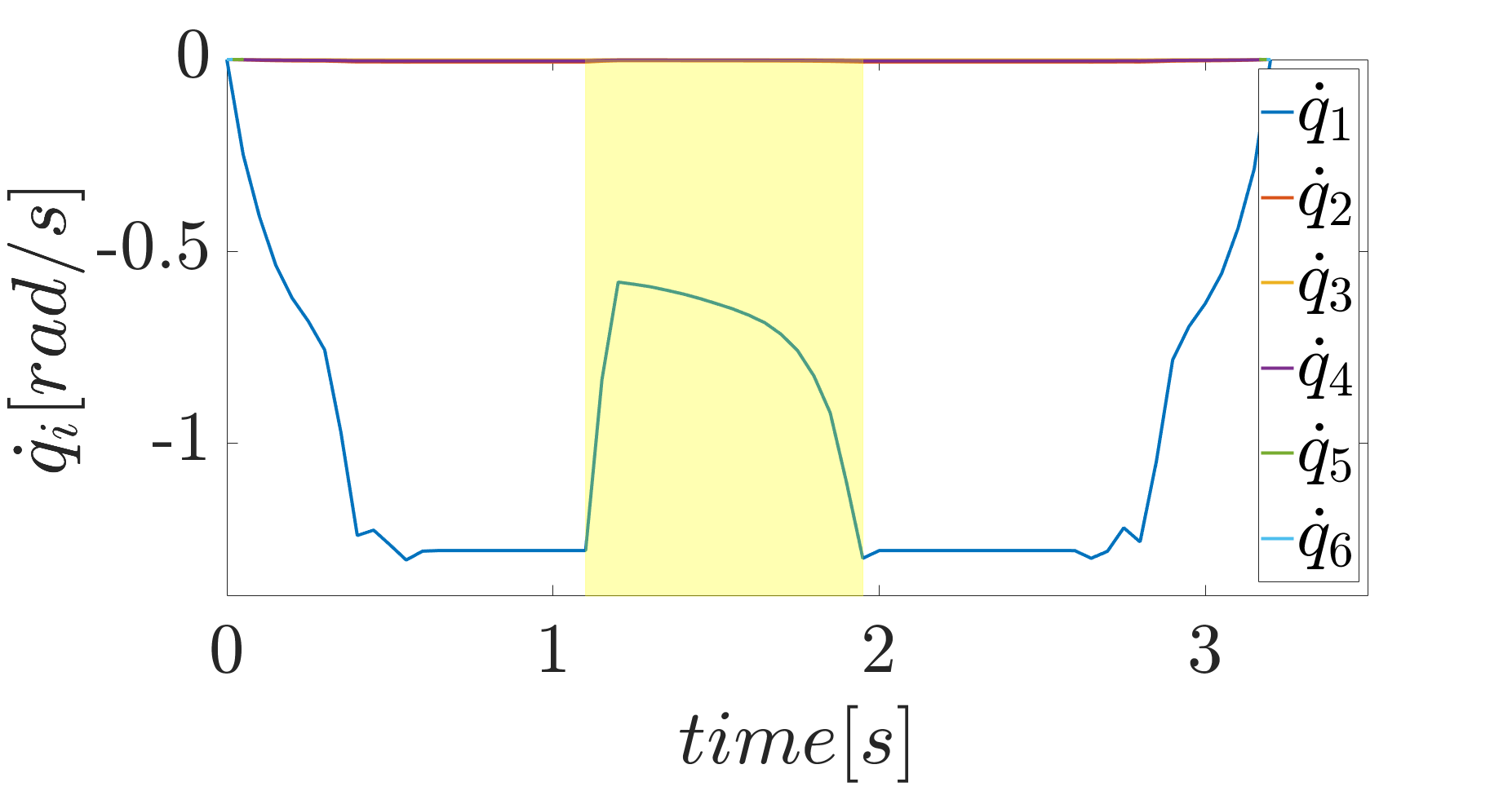}} 
    \caption{Second part of the experiment. As a consequence of the scaling, the robot requires more time to complete the trajectory.}
    \label{fig:slow}
\end{figure}

In the experiment the robot has to go continually from the start configuration \mbox{$q_s = \{1.57, -0.4, 1.17, 0.0, 1.57, 0.0\}$} to the final configuration $q_g =\{-1.57, -0.4, 1.17, 0.0, 1.57, 0.0\}$, and vice versa. A complete video of the demonstration is attached.  Initially, the human operator is very far from the robot, i.e. he is in the green area of the SSM (see Fig.~\ref{fig:ssm}). In this phase the robot is allowed to move at maximum speed, following the nominal planned trajectory as shown in Fig.~\ref{fig:nominal}.
Subsequently, the human operator approaches the robot causing the scaling of the trajectory, as shown in Fig.~\ref{fig:scale}. This is due to the fact that, according to the safety limit imposed by ISO/TS 15066 \eqref{eq:vellimit}, the maximum speed allowed towards the human operator decreases. The Fig.~\ref{fig:human_pose} and \ref{fig:human_vel} show the position and the velocity of the nearest human point in the robot reference frame, respectively. As a consequence of the approaching behavior, in the first phase the $x$ component increases and its velocity is positive. While during the scaling, the velocity components are very low. The Fig~\ref{fig:velconst} demonstrates that the safety constraint is not violated. In the graph only the velocity of the end-effector towards the human $v_{ee}^H$ is shown. It is worth noting that the robot slows down only in the first part of the trajectory, i.e. from $t=1.1\,\,sec$ to $t=1.95\,\,sec$. This is because the robot is going towards the human operator. In the second part, i.e. when it moves away, it goes at higher speed, restoring the nominal behavior. As a matter of fact, at $t=1.95\,\,sec$ the scaling factor increases. A comparison between the planned trajectory and the scaled one can be found in Fig~\ref{fig:slow}.

 In the next part of the experiment the human operator hinders the robot, making the trajectory infeasible. The dynamic planner layer takes care of planning a new one, avoiding the human operator, and the robot is able to reach the desired configuration. 
 
\begin{figure}[t]
	\centerline{\includegraphics[width=0.8\columnwidth]{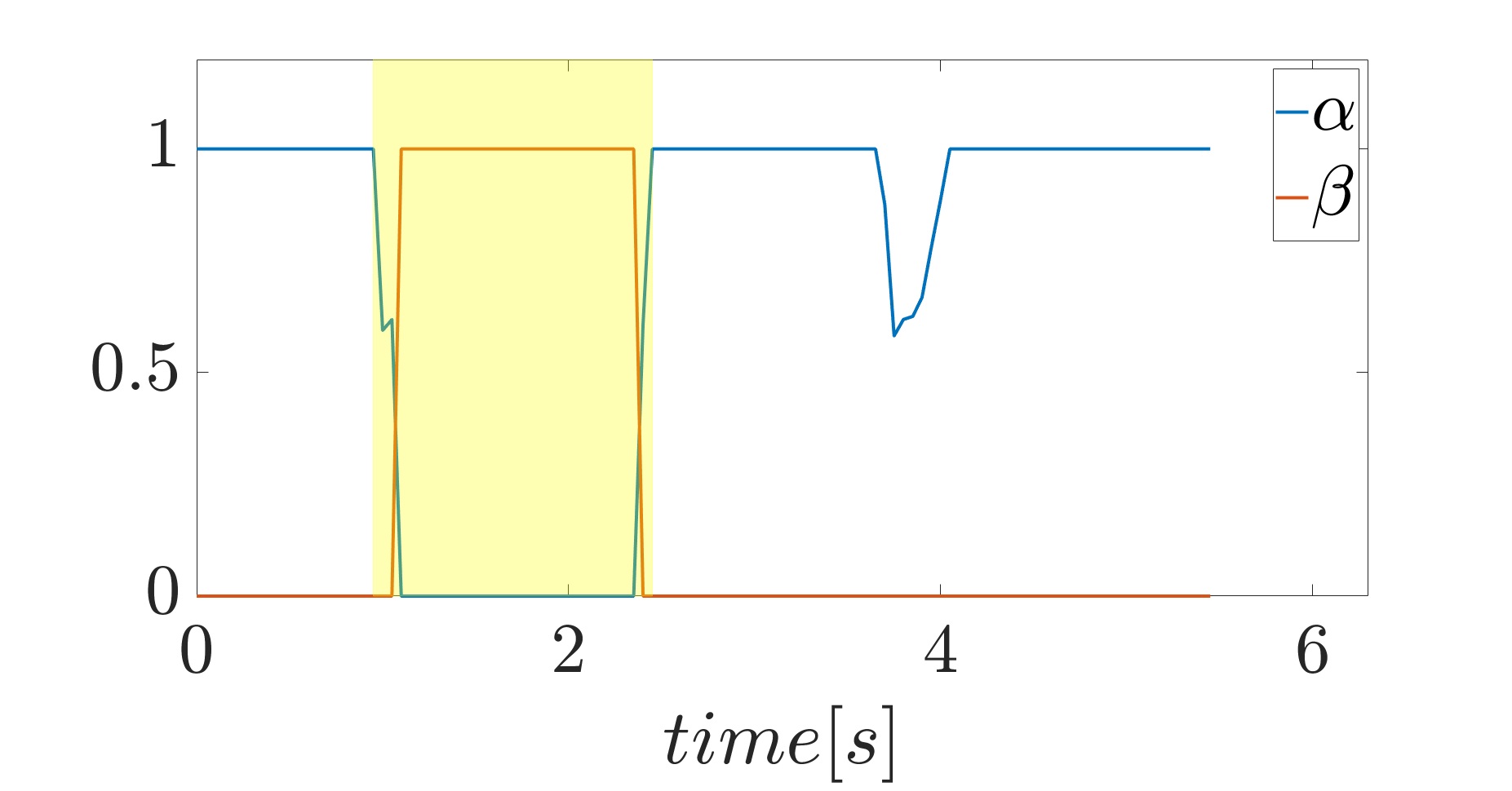}}
	\caption{Last part of the experiment. The scaling factor drops lower the threshold, the step signal is activated and the replanning request is sent.} 
	\label{fig:slow_scal_rep}
\end{figure}

 In the last part of the experiment, the human operator goes very close to the robot, causing a drop of the scaling factor. When $\alpha\le\alpha_{min}=0.2$ the trajectory scaling layer sends a the step signal $\beta=1$ to the dynamic planner requesting for a replan of a more efficient trajectory, as explained in Sec.~\ref{sec:architecture}. The evolution of the scaling factor and the signal is shown in Fig.~\ref{fig:slow_scal_rep}.
The replanning strategy is successful and the robot is free to restore its behavior, completing the trajectory. Please note that the robot remains stopped for about $1.5\,\, sec$. This is due to the fact that the dynamic planner requires a certain amount of time to find the more efficient trajectory.

In order to demonstrate the effectiveness of the architecture, the same experiment is performed without sending the replan request when the scaling factor is too small. In the last part of the video is shown that, when the human operator goes very close, the robot stops and it stays stuck until the human operator leaves. It is worth noting that a comparison on the execution times of the different strategies would not be very interesting. This is because the solution without the replan signal strictly depends on how long the human operator stays close to the robot.
\section{Conclusions} 
\label{sec:conclusions}
In this paper we propose a two-layers framework for trajectory planning and velocity scaling. Taking into account the human motion, the first layer, i.e. the dynamic planner layer, continuously checks if the trajectory becomes infeasible and and reacts accordingly. The second layer, i.e. the trajectory scaling layer, explicitly considers the safety standards and scales the robot velocity in order to ensure safety. Moreover, when the scaling factor decreases too much, the trajectory scaling sends a signal to the dynamic planner requesting for a replanning of a new trajectory. The experimental evaluation shows the effectiveness of the framework both when the human operator hinders the robot and when the two agents get too close.

Future work aims to exploit the model predictive control approach to generate smoother speed profiles. Furthermore, considering a single direction from each robot link to the human operator can be quite unreliable from a safety point of view. For this reason a strategy that considers multiple directions could be implemented. This, will be tested in a real scenario with the presence of many obstacles, e.g. a cluttered environment. Finally, the work could be extended for use also in the case of mobile manipulators, exploiting the redundancy to better improve the safety.
\bibliographystyle{IEEEtran}
\bibliography{bibliography}

\begin{thebibliography}{10}
\providecommand{\url}[1]{#1}
\csname url@samestyle\endcsname
\providecommand{\newblock}{\relax}
\providecommand{\bibinfo}[2]{#2}
\providecommand{\BIBentrySTDinterwordspacing}{\spaceskip=0pt\relax}
\providecommand{\BIBentryALTinterwordstretchfactor}{4}
\providecommand{\BIBentryALTinterwordspacing}{\spaceskip=\fontdimen2\font plus
\BIBentryALTinterwordstretchfactor\fontdimen3\font minus
  \fontdimen4\font\relax}
\providecommand{\BIBforeignlanguage}[2]{{%
\expandafter\ifx\csname l@#1\endcsname\relax
\typeout{** WARNING: IEEEtran.bst: No hyphenation pattern has been}%
\typeout{** loaded for the language `#1'. Using the pattern for}%
\typeout{** the default language instead.}%
\else
\language=\csname l@#1\endcsname
\fi
#2}}
\providecommand{\BIBdecl}{\relax}
\BIBdecl

\bibitem{villani2018}
V.~Villani, F.~Pini, F.~Leali, and C.~Secchi, ``Survey on human--robot
  collaboration in industrial settings: Safety, intuitive interfaces and
  applications,'' \emph{Mechatronics}, vol.~55, pp. 248--266, 2018.

\bibitem{iso2011robot-1}
``{Robots and Robotic Devices--Safety Requirements for Industrial Robots--Part
  1: Robots},'' International Organization for Standardization, Geneva, CH,
  Standard, Jul. 2011.

\bibitem{iso2011robot-2}
``{Robots and Robotic Devices--Safety Requirements for Industrial Robots--Part
  2: Robot systems and integration},'' International Organization for
  Standardization, Geneva, CH, Standard, Jul. 2011.

\bibitem{isots}
``{Robots and robotic devices--Collaborative robots},'' International
  Organization for Standardization, Geneva, CH, Technical Specification, Feb.
  2016.

\bibitem{ragaglia2015}
M.~Ragaglia, A.~M. Zanchettin, and P.~Rocco, ``Safety-aware trajectory scaling
  for human-robot collaboration with prediction of human occupancy,'' in
  \emph{2015 International Conference on Advanced Robotics (ICAR)}.\hskip 1em
  plus 0.5em minus 0.4em\relax IEEE, 2015, pp. 85--90.

\bibitem{zanchettin2015}
A.~M. Zanchettin, N.~M. Ceriani, P.~Rocco, H.~Ding, and B.~Matthias, ``Safety
  in human-robot collaborative manufacturing environments: Metrics and
  control,'' \emph{IEEE Transactions on Automation Science and Engineering},
  vol.~13, no.~2, pp. 882--893, 2015.

\bibitem{lippi2020}
M.~Lippi and A.~Marino, ``Human multi-robot safe interaction: A trajectory
  scaling approach based on safety assessment,'' \emph{IEEE Transactions on
  Control Systems Technology}, 2020.

\bibitem{levratti2019}
A.~Levratti, G.~Riggio, C.~Fantuzzi, A.~De~Vuono, and C.~Secchi, ``Tirebot: A
  collaborative robot for the tire workshop,'' \emph{Robotics and
  Computer-Integrated Manufacturing}, vol.~57, pp. 129--137, 2019.

\bibitem{chen2018}
J.-H. Chen and K.-T. Song, ``Collision-free motion planning for human-robot
  collaborative safety under cartesian constraint,'' in \emph{2018 IEEE
  International Conference on Robotics and Automation (ICRA)}.\hskip 1em plus
  0.5em minus 0.4em\relax IEEE, 2018, pp. 1--7.

\bibitem{ferraguti2015}
F.~Ferraguti, N.~Preda, M.~Bonfe, and C.~Secchi, ``Bilateral teleoperation of a
  dual arms surgical robot with passive virtual fixtures generation,'' in
  \emph{2015 IEEE/RSJ International Conference on Intelligent Robots and
  Systems (IROS)}.\hskip 1em plus 0.5em minus 0.4em\relax IEEE, 2015, pp.
  4223--4228.

\bibitem{lin2017}
H.-C. Lin, C.~Liu, Y.~Fan, and M.~Tomizuka, ``Real-time collision avoidance
  algorithm on industrial manipulators,'' in \emph{2017 IEEE Conference on
  Control Technology and Applications (CCTA)}.\hskip 1em plus 0.5em minus
  0.4em\relax IEEE, 2017, pp. 1294--1299.

\bibitem{ferraguti2020-iso}
F.~Ferraguti, M.~Bertuletti, C.~T. Landi, M.~Bonf{\`e}, C.~Fantuzzi, and
  C.~Secchi, ``A control barrier function approach for maximizing performance
  while fulfilling to iso/ts 15066 regulations,'' \emph{IEEE Robotics and
  Automation Letters}, vol.~5, no.~4, pp. 5921--5928, 2020.

\bibitem{ames2016}
A.~D. Ames, X.~Xu, J.~W. Grizzle, and P.~Tabuada, ``Control barrier function
  based quadratic programs for safety critical systems,'' \emph{IEEE
  Transactions on Automatic Control}, vol.~62, no.~8, pp. 3861--3876, 2016.

\bibitem{lavalle2001}
S.~M. LaValle and J.~J. Kuffner~Jr, ``Randomized kinodynamic planning,''
  \emph{The international journal of robotics research}, vol.~20, no.~5, pp.
  378--400, 2001.

\bibitem{kunz2014}
T.~Kunz and M.~Stilman, ``Probabilistically complete kinodynamic planning for
  robot manipulators with acceleration limits,'' in \emph{2014 IEEE/RSJ
  International Conference on Intelligent Robots and Systems}.\hskip 1em plus
  0.5em minus 0.4em\relax IEEE, 2014, pp. 3713--3719.

\bibitem{udai2014}
A.~D. Udai, A.~A. Hayat, and S.~K. Saha, ``Parallel active/passive force
  control of industrial robots with joint compliance,'' in \emph{2014 IEEE/RSJ
  International Conference on Intelligent Robots and Systems}.\hskip 1em plus
  0.5em minus 0.4em\relax IEEE, 2014, pp. 4511--4516.

\bibitem{moon2016}
S.~Moon, Y.~Park, D.~W. Ko, and I.~H. Suh, ``Multiple kinect sensor fusion for
  human skeleton tracking using kalman filtering,'' \emph{International Journal
  of Advanced Robotic Systems}, vol.~13, no.~2, p.~65, 2016.

\bibitem{kofman2005}
J.~Kofman, X.~Wu, T.~J. Luu, and S.~Verma, ``Teleoperation of a robot
  manipulator using a vision-based human-robot interface,'' \emph{IEEE
  transactions on industrial electronics}, vol.~52, no.~5, pp. 1206--1219,
  2005.

\bibitem{fan2010}
J.~Fan, W.~Xu, Y.~Wu, and Y.~Gong, ``Human tracking using convolutional neural
  networks,'' \emph{IEEE Transactions on Neural Networks}, vol.~21, no.~10, pp.
  1610--1623, 2010.

\bibitem{lavalle1998}
S.~M. LaValle, ``Rapidly-exploring random trees: A new tool for path
  planning,'' 1998.

\bibitem{jaillet2004}
L.~Jaillet and T.~Sim{\'e}on, ``A prm-based motion planner for dynamically
  changing environments,'' in \emph{2004 IEEE/RSJ International Conference on
  Intelligent Robots and Systems (IROS)(IEEE Cat. No. 04CH37566)},
  vol.~2.\hskip 1em plus 0.5em minus 0.4em\relax IEEE, 2004, pp. 1606--1611.

\bibitem{ratliff2009}
N.~Ratliff, M.~Zucker, J.~A. Bagnell, and S.~Srinivasa, ``Chomp: Gradient
  optimization techniques for efficient motion planning,'' in \emph{2009 IEEE
  International Conference on Robotics and Automation}.\hskip 1em plus 0.5em
  minus 0.4em\relax IEEE, 2009, pp. 489--494.

\bibitem{ferraguti2020}
F.~Ferraguti, C.~T. Landi, S.~Costi, M.~Bonf{\`e}, S.~Farsoni, C.~Secchi, and
  C.~Fantuzzi, ``Safety barrier functions and multi-camera tracking for
  human--robot shared environment,'' \emph{Robotics and Autonomous Systems},
  vol. 124, p. 103388, 2020.

\bibitem{optitrack}
\BIBentryALTinterwordspacing
NaturalPoint. (2020) Optitrack - motion capture systems. [Online]. Available:
  \url{https://www.optitrack.com/}
\BIBentrySTDinterwordspacing

\bibitem{rrt-connect}
J.~J. Kuffner and S.~M. LaValle, ``Rrt-connect: An efficient approach to
  single-query path planning,'' in \emph{Proceedings 2000 ICRA. Millennium
  Conference. IEEE International Conference on Robotics and Automation.
  Symposia Proceedings (Cat. No. 00CH37065)}, vol.~2.\hskip 1em plus 0.5em
  minus 0.4em\relax IEEE, 2000, pp. 995--1001.

\bibitem{moveit}
D.~Coleman, I.~Sucan, S.~Chitta, and N.~Correll, ``Reducing the barrier to
  entry of complex robotic software: a moveit! case study,'' \emph{arXiv
  preprint arXiv:1404.3785}, 2014.

\bibitem{cvxgen}
J.~Mattingley and S.~Boyd, ``{CVXGEN}: A code generator for embedded convex
  optimization,'' \emph{Optimization and Engineering}, vol.~12, no.~1, pp.
  1--27, 2012.

\end{thebibliography}
 \end{document}